\pdfoutput=1

\documentclass[11pt]{article}

\usepackage[final]{acl}
\usepackage{tcolorbox}
\usepackage{pifont}  
\usepackage{wrapfig}
\usepackage{float}
\usepackage{placeins} 
\usepackage{tabularx} 
\usepackage{pdfpages}
\usepackage{needspace}
\usepackage{titlesec}
\usepackage{listings}
\usepackage[dvipsnames]{xcolor}

\tcbuselibrary{breakable,skins,listingsutf8}

\usepackage{times}
\usepackage{latexsym}

\usepackage{amsmath} 
\usepackage[T1]{fontenc}

\usepackage[utf8]{inputenc}
\usepackage{booktabs}
\usepackage{siunitx}
\usepackage{microtype}

\usepackage{inconsolata}

\usepackage{graphicx}
\usepackage{courier}
\usepackage{amsmath}
\usepackage{multirow}
\usepackage{multicol}
\usepackage{colortbl}
\usepackage{xcolor}  
\usepackage{subcaption}
\usepackage{float}  
\usepackage{afterpage}
\usepackage{caption}
\usepackage{array}
\usepackage{makecell}
\usepackage{placeins}  
\usepackage{afterpage}  
\usepackage{enumitem}

\title{Template-Based Text-to-Image Alignment for Language Accessibility: A Study on Visualizing Text Simplifications}

\newcommand{\uzh}{\includegraphics[width=1.5em]{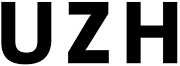}}

\author{
\hspace{1em}
  \textbf{Belkiss Souayed}$^{\uzh}$ \hspace{0.5em}
  \textbf{Sarah Ebling}$^{\uzh}$ \hspace{0.5em}
  \textbf{Yingqiang Gao\footnotemark[2]}$^{\uzh}$
  \\
  \hspace{1em}
  $^{\uzh}$Department of Computational Linguistics, University of Zurich, Switzerland \\
  \hspace{0.6em}
  \texttt{belkiss.souayed@uzh.ch} \\
  \texttt{\{ebling, yingqiang.gao\}@cl.uzh.ch} \\
}

\renewcommand{\thefootnote}{\fnsymbol{footnote}}

\begin{document}
\maketitle

\maketitle
\begin{center}
\begin{minipage}{\textwidth}
\vspace{-6.5em}
\centering
{\color{red}\bfseries
Warning: This paper includes AI-generated images that may cause visual discomfort.
}
\vspace{0.8em}
\end{minipage}
\end{center}

\footnotetext[2]{Corresponding author.}

\renewcommand{\thefootnote}{\arabic{footnote}}

\begin{abstract}

Individuals with intellectual disabilities often have difficulties in comprehending complex texts. While many text-to-image models prioritize aesthetics over accessibility, it is not clear how visual illustrations relate to text simplifications (TS) generated from them. This paper presents a structured vision-language model (VLM) prompting framework for generating accessible images from simplified texts. We designed five prompt templates, i.e., \textit{Basic Object Focus}, \textit{Contextual Scene}, \textit{Educational Layout}, \textit{Multi-Level Detail}, and \textit{Grid Layout}, each following distinct spatial arrangements while adhering to accessibility constraints such as object count limits, spatial separation, and content restrictions. Using 400 sentence-level simplifications from four established TS datasets (OneStopEnglish, SimPA, Wikipedia, and ASSET), we conducted a two-phase evaluation: Phase 1 assessed prompt template effectiveness with CLIPScores, and Phase 2 involved human annotation of generated images across ten visual styles by four accessibility experts. Results show that the \textit{Basic Object Focus} prompt template achieved the highest semantic alignment, indicating that visual minimalism enhances language accessibility. Expert evaluation further identified \textit{Retro} style as the most accessible and Wikipedia as the most effective data source. Inter-annotator agreement varied across dimensions, with \textit{Text Simplicity} showing strong reliability and \textit{Image Quality} proving more subjective. Overall, our framework offers practical guidelines for accessible content generation and underscores the importance of structured prompting in AI-generated visual accessibility tools.

\noindent
\begin{minipage}{\columnwidth}
\centering
\raisebox{-0.1cm}{\hspace{-4em}\includegraphics[height=0.5cm]{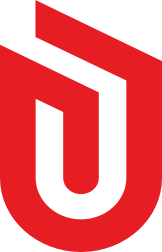}}~
\href{https://www.swissubase.ch/en/share/eCTXtvik5Lr96iLcAk0nfnfSheu-QlACjK6TT-jZ6Q8c_ZOYu3VZfDaVF1AW9jsRRL8}{\textcolor{gray}{Dataset}} \quad
\raisebox{-0.15cm}{\includegraphics[height=0.5cm]{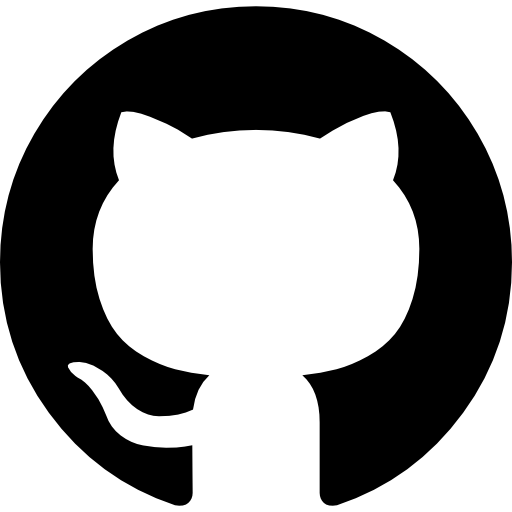}}~
\href{https://github.com/ZurichNLP/for-belkiss-souayed/tree/main}{\textcolor{gray}{Code}}
\end{minipage}

\end{abstract}

\section{Introduction}

Individuals with intellectual disabilities often have difficulties understanding complex texts \citep{alva2020asset, yawiloeng2022using}. While text simplification (TS) improves readability, it is frequently insufficient on its own. Research shows that visual support, recommended by Easy-to-Read guidelines, can significantly enhance comprehension \citep{madina2023easy}. However, most text-to-image models prioritize aesthetics over accessibility, which can cause cluttered, abstract, or semantically misaligned visuals generated from text inputs.

\begin{figure}[t]
  \centering
  \includegraphics[width=0.4\textwidth]{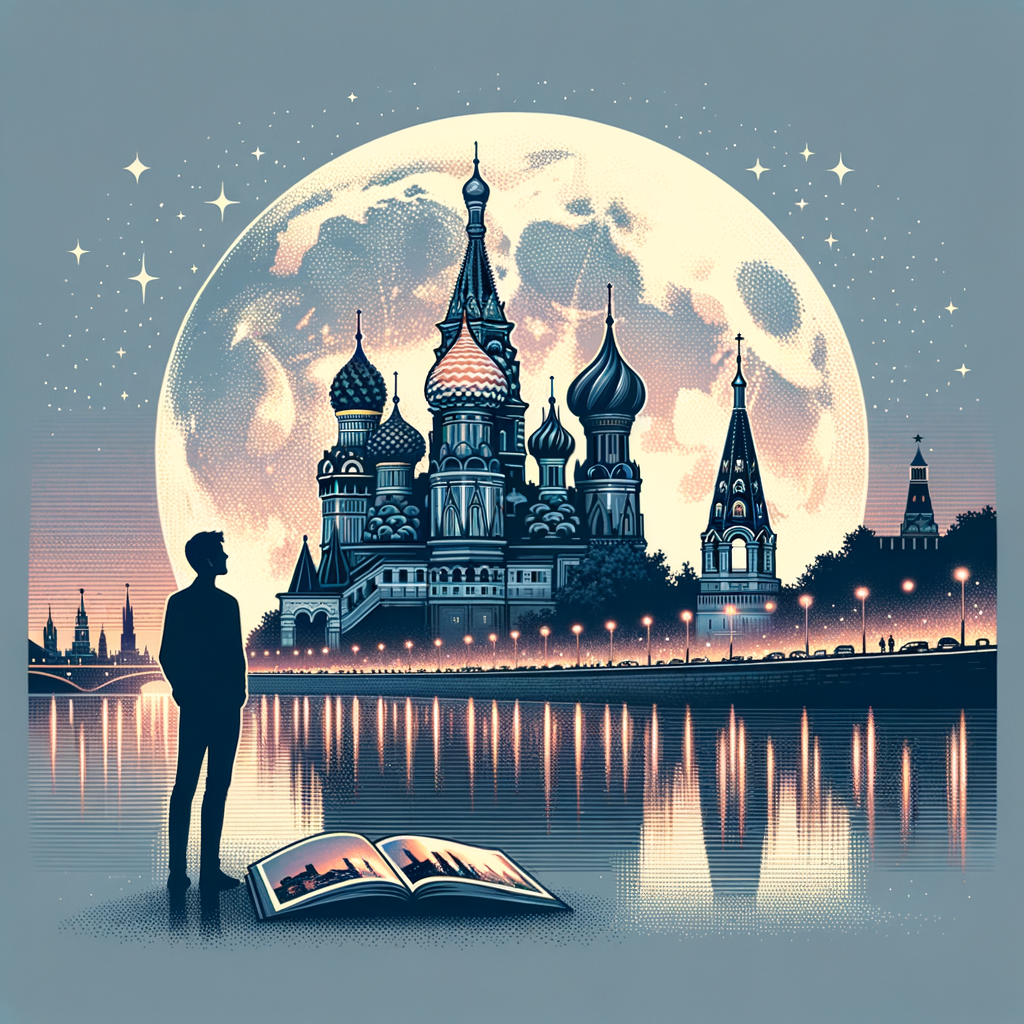}
  \caption{Example image generated based on the simplified text 
  \emph{``I will never forget the wonderful memories he has given us, like that magical night in Moscow.''} 
  (\textbf{Style:} Artistic, \textbf{Dataset:} OneStopEnglish).}
  \label{fig:illustrative}
\end{figure}

Recent studies have explored image retrieval for simplified text as a potential solution \citep{geislinger2023multi, singh2023enhancing}. However, none have systematically investigated structured prompting for accessible image generation in the TS context. In this work, we address this gap by introducing a template-based framework that explicitly enforces accessibility constraints, such as controlled object count, spatial separation, and the absence of embedded text, directly within prompts for querying vision-language models (VLMs).

Using 400 text simplification pairs from four datasets (ASSET, OneStopEnglish, SimPA, and Wikipedia), we generated and evaluated 4,000 images across five prompt templates and ten visual styles. Our study aims to answer the following research questions (RQs):
\begin{itemize}[noitemsep, topsep=0pt, partopsep=0pt, parsep=0pt, left=0pt]
\item \textbf{RQ1}: How can template-based prompting improve the accessibility of images generated from simplified texts?
\item \textbf{RQ2}: Which visual styles and data sources are most effective for accessible image generation?
\item \textbf{RQ3}: How do expert annotations compare to automatic evaluation metrics in assessing accessibility?
\end{itemize}

Our results show that the \textit{Basic Object Focus} template performs best, supporting visual minimalism. Experts rate \textit{Retro} style and \textit{Wikipedia}-sourced simplifications as most accessible. We also find weak correlation between CLIPScores and human judgments, underscoring the need for human-centered evaluation in accessible and inclusive AI.

\section{Related Work}
\label{sec:relatedwork}

\subsection{Visual-Aided Text Simplification}

As a natural language processing (NLP) task to improve language accessibility, TS actively modifies complex texts to improve readability for diverse target group persons, including persons with cognitive disabilities \citep{espinosa2023automatic, gao2025evaluting}, foreign language learners \citep{degraeuwe2022lexical, li2025aligning} and others. TS can operate at the lexical, syntactic, and discourse levels \citep{alva2020asset, zhong2020discourse}. 
Beyond language, accessible communication increasingly benefits from multimodal aids, since simplified text alone may not suffice for users who have reading difficulties. Studies show that pairing text with visual components improves overall comprehension and engagement \citep{yawiloeng2022using} and mirrors those principles in Easy-to-Read guidelines \citep{madina2023easy}, which recommend simple text with visual illustrations for better language accessibility. However, producing accessible visuals remains a technical challenge, motivating research interest in automatic image retrieval and text-to-image generation in a TS context \citep{marturi2025llm}. Our study builds on this line of work by investigating how  images generated through VLMs from text simplifications can support language accessibility through both semantic simplicity and visual clarity.

\subsection{Retrieval-based Language Accessibility}

Early multimodal approaches retrieve images from databases to support the language accessibility. \citet{geislinger2023multi} combined NLP approaches with eye-tracking to provide real-time visual support for difficult terms, while \citet{singh2023enhancing} optimized textbook enrichment by retrieving web images using CLIP-based similarity \citep{radford2021learning}. Such systems improved perceived educational outcomes, but they remain limited by the availability of accessible visuals in existing databases, especially for abstract or personalized concepts. These challenges further motivate research in text-to-image alignment for language accessibility. 

\subsection{Datasets and Text-to-Image Benchmarks}

Recent advances in VLMs such as CLIP \citep{radford2021learning} and DALL·E \citep{betker2023improving} have driven progress in text-to-image alignment studies. Several datasets have been proposed to either train image captioning models or evaluate the alignment quality.

Benchmark datasets like MS COCO \citep{lin2014microsoft} and Flickr30K \citep{plummer2015flickr30k} focus on short image captions but are not specific to accessibility for simplified text. MOTIF \citep{wang2022motif} pairs simplified sentences with illustrative images to support second-language learners, though it remains limited to multimodal retrieval rather than generation. Our work addresses this gap by generating new images directly from text simplifications, enabling scalable visual support.

\citet{anschutz2024images} present the first systematic study of text-to-image for Easy-to-Read German content. Using 80 structured prompts translated from German, authors generated 2,240 images across seven models and evaluated them with automated metrics FID \citep{heusel2017gans}, CLIPScore  \citep{radford2021learning}, and TIFA  \citep{hu2023tifa}. 

While previous works have investigated text-to-image alignment in the context of image captioning and alignment evaluation, to the best of our knowledge, no prior work has proposed a structured prompting strategy that explicitly enforces accessibility constraints such as controlled object count, spatial separation, and the exclusion of abstract or textual elements. Likewise, no previous study has compared the accessibility effects of different visual styles in a systematic way. In this work, we address these research gaps by introducing a template-based prompting framework aligned with accessibility principles and conducting a large-scale experiment: 4,000 images generated from 400 simplified sentences, spanning five prompt templates and ten visual styles. In contrast to fixed-format prompts, our pipeline transforms naturally simplified sentences into structured prompts while preserving semantic content. We combine automatic metrics with expert annotations across multiple dimensions, enabling the first human evaluation of how prompt design and visual style influence cognitively accessible image generation.

\section{Methodology}
\label{sec:methodology}

Our study follows a two-phase pipeline. In \textbf{Phase 1}, we developed and evaluated five prompt templates used by VLMs for generating visuals, each implementing unique accessibility constraints while sharing core principles such as:
\begin{itemize}[noitemsep, topsep=0pt, partopsep=0pt, parsep=0pt, left=0pt]
    \item \textbf{Controlled object counts}: Defining the quantity of visual components present in the generated images;
    \item \textbf{Spatial separation}: Checking if visual components are physically well-arranged;
    \item \textbf{Exclusion of text or abstract elements}: Examining whether VLMs render texts in generated images, which can cause additional confusion.
\end{itemize}

In \textbf{Phase 2}, we take the best-performing template and apply it to produce a large-scale multimodal dataset. We generated 4,000 images from 400 simplified sentences, spanning ten visual styles, with style-specific prompt adjustments  (see Appendix~\ref{app:prompt-example} for an illustration of how a single sentence generates multiple style-specific prompts). Four expert annotators then assessed the resulting text–image pairs across six dimensions using pre-defined criteria designed to capture both quantitative and qualitative accessibility aspects. 

While we acknowledge that images are linked to textual units of varying granularity, operating on the sentence level throughout was a means of  isolating the effect of text unit size.

\subsection{Prompt Templates}

Our initial trials with direct prompts (e.g., \textit{``Generate an accessible image for this simplified sentence: [sent]''}) produced generic, inconsistent results. Without structural guidance, the model often failed to enforce key accessibility needs such as clear object focus, reduced background clutter, or stylistic appropriateness.
Although the templates were structurally pre-defined, \texttt{GPT-4} \citep{achiam2023gpt} was necessary to map sentence semantics into visual prompts. It interpreted each simplified sentence to decide which objects and relations to depict, how to represent abstract ideas concretely, and which visual style to apply. This semantic-to-visual prompt generation goes beyond simple template filling or rule-based rewriting.
We concluded that such direct prompts lacked the precision needed to control the layout structure and enforce necessary visual requirements such as selecting the most suitable stylistic representation, maintaining a clear depiction of primary objects without clutter, and minimizing background noise to reduce cognitive load. 

To address this, we adopted structured templates that embed accessibility constraints directly into the prompt. Template design was grounded in W3C Web Accessibility Initiative (WAI) image tutorials\footnote{W3C Web Accessibility Initiative (WAI) image tutorials: guidance on \textit{informative images}, \textit{complex images}, and \textit{groups of images}. Available at: \url{https://www.w3.org/WAI/tutorials/images/}}, which classify accessible images as \textit{informative} (conveying concepts visually), \textit{complex} (layered diagrams), or \textit{grouped} (collections representing unified information). Each template isolates a specific layout logic, e.g., alignment, sequencing, or hierarchy, to assess its impact on visual comprehension, reflecting formats common in educational and assistive contexts:
\begin{itemize}[noitemsep, topsep=0pt, partopsep=0pt, parsep=0pt, left=0pt]
    \item \textbf{Basic Object Focus:} Removes all spatial context to measure the effect of object isolation. Isolates between one and three objects on a plain background to maximize clarity and minimize cognitive load;
    \item \textbf{Contextual Scene:} Presents simple real-world layouts (e.g., items on a shelf) to test how minimal grounding supports understanding. Situates objects in simple real-world settings (e.g., items in a room) to provide minimal contextual grounding;
    \item \textbf{Educational Layout:} Introduces sequencing and flow, mimicking instructional visuals like timelines. Uses simple relations such as arrows to support instructional use;
    \item \textbf{Multi-Level Detail:} Inspired by textbook diagrams, adds foreground–background layering to explore hierarchy and layered perception. Layers a central object with two to three related subobjects to explore hierarchical perception;
    \item \textbf{Grid Layout:} Simulates classification interfaces (e.g., icon grids) to examine whether symmetry aids clarity. Organizes items in a 2×2 or 3×3 structure, simulating classification interfaces and testing whether symmetry aids clarity.
\end{itemize}

All five templates share a unified set of accessibility constraints (see Appendix~\ref{app:prompt-templates} for detailed specifications). 
To ensure fair comparison and accessibility, the generation process enforced baseline constraints: 
(1) between three to five distinct objects per image; 
(2) sufficient spatial separation between objects; 
(3) avoidance of text, numbers, or motion effects; 
(4) exclusion of abstract, metaphorical, or culturally biased elements; 
and (5) preference for plain or neutral backgrounds. 
These requirements operationalize accessibility principles and reduce potential bias.

To identify which prompt template best supports cognitively accessible image generation, we generated 100 images per template (500 total) from simplified sentences. Prompts were constructed with \texttt{GPT-4} \citep{achiam2023gpt} and images generated using \texttt{DALL·E 3} \citep{betker2023improving}. Outputs were assessed using CLIPScore (\texttt{ViT-L/14@336px}; \citet{hessel2021clipscore}) to measure text-to-image alignment. We chose these models for their high prompt fidelity, built-in safety filters, and consistency across large batches. As the focus was on evaluating prompt design rather than comparing VLMs or model-specific optimization, alternative architectures were not considered.

To guide template selection, we developed a composite scoring system to balance accuracy and robustness rather than optimize a single metric. Weights were empirically chosen to prevent unstable templates from dominating, ensuring selection of templates that perform reliably across diverse simplification contexts rather than excelling on a narrow subset of inputs. Given a template $t$, the composite score is computed as:
\begin{equation*}
\resizebox{\columnwidth}{!}{$
\begin{aligned}
& \text{Composite}(t) \\
&= 0.4 \mu_t + 0.2 C_t + 0.2 S_t + 0.1 B_t + 0.1 (1 - W_t),
\end{aligned}
$}
\end{equation*}
where $\mu_t$ is the mean CLIPScore (40\%), $C_t$ is the consistency (20\%), $S_t$ is the success rate (20\%), 
$B_t$ is the fraction of best-performing cases (10\%), and $W_t$ is the fraction of worst-performing cases (10\%). 
All components were normalized to $[0,1]$ before aggregation.

\begin{table}[htbp]
    \centering
    \resizebox{0.8\columnwidth}{!}{
    \begin{tabular}{lcc}
        \toprule
        \textbf{Template} & \textbf{CLIP} & \textbf{Composite} \\
        \midrule
        \rowcolor{gray!20!white}
        Basic Object Focus   & \textbf{0.211} & \textbf{5.31} \\
        Contextual Scene     & 0.210 & 4.95 \\
        Educational Layout   & 0.202 & 4.63 \\
        Multi-Level Detail   & 0.201 & 4.50 \\
        Grid Layout          & 0.199 & 4.39 \\
        \bottomrule
    \end{tabular}
    }
    \caption{Phase~1 results (100 prompts × 5 templates). Basic Object Focus ranked highest and was selected for Phase~2.}
    \label{tab:clip_template_results}
\end{table}

As shown in Table~\ref{tab:clip_template_results}, the \textit{Basic Object Focus} template achieved the highest text-to-image alignment, consistency, and overall composite score. This suggests that reducing background clutter and emphasizing a small number of core objects is most effective for supporting accessibility. Based on these findings, \textit{Basic Object Focus} was selected as the foundation for Phase~2, where we scaled generation to 4,000 images across ten visual styles and conduct expert evaluations to address RQ2 and RQ3.

\subsection{Dataset Compilation}
\label{sec:dataset}

We compiled a text-to-image TS corpus by sampling from four established text simplification datasets: OneStopEnglish \citep{vajjala2018onestopenglish}, SimPA \citep{scarton2018simpa}, Wikipedia (\citet{sun2020helpfulness}, without context), and ASSET \citep{alva2020asset}. These corpora were selected for their complementary domains and simplification strategies, covering news, public administration, encyclopedic text, and web content, each provides parallel complex–simplified sentence pairs. Table~\ref{tab:ch3_datasets} in Appendix~\ref{app:expert-examples} summarizes their main features.

The four source datasets are all sentence-aligned, offering multiple simplifications per complex text or varying types of simplification methods. This led to the necessity of adopting a consistent sampling strategy to support uniformity and methodological coherence across all datasets. Specifically, among the multiple simplifications available for each complex sentence, we retained only one per original to ensure a one-to-one mapping between the simplification and its visually generated counterpart. We also ensured to randomly draw the exact number of samples from each corpus in order to create a balanced dataset which  reflects the distinct simplification approaches and domains covered within each source text. 




We sampled a balanced subset of 400 pairs (100 per dataset). The sampling procedure used random selection to avoid bias and ensure that sampled instances reflected the overall linguistic variability and quality of the source datasets. We deliberately avoided cherry-picking instances we believed to be ideal for generation or applying quality filtering. By ``ideal'' or ``quality'', we refer to sentences with simple syntax and clear meaning as opposed to abstract or ambiguous ones that are harder to visualize. This approach ensured our evaluation methodology reflects those real-world difficulties of accessible image generation across diverse text features.

The  400 sentence pairs were stored in a \texttt{JSON} format, where each line represents one structured data entry. In addition to the complex and its simplified counterpart, we annotated each pair with relevant metadata fields aimed to support traceability and data hierarchy (see Appendix~\ref{app:datasets} for a complete example). This dataset served as the basis for prompt formulation, image generation, and expert annotation. 

As all TS datasets were human-produced and sentence-aligned, we conducted minimal data preprocessing including removal of formatting artifacts (particularly in Wikipedia) and standardizing all samples into \texttt{JSON} format with metadata for dataset source, domain, and token counts. We focused on filtering overly short (<10 tokens) or long (>55 tokens) sentence-level simplifications to ensure balanced text complexity. After pre-processing, the final dataset exhibited an average complex sentence length of 26.2 tokens and simplified sentence length of 23.9, with an overall reduction of 2.3 tokens (8.8\%). 

Following Phase~1, we refined \textit{Basic Object Focus} to preserve minimalism while improving reliability: exactly four objects, at least 30\% spacing between objects, and a 10\% cap on size variation to maintain equal prominence. A pilot trial on 20 samples yielded a CLIPScore of 0.3465, a 64\% increase over the Phase~1 score (0.2108), confirming the benefit of explicit spacing and uniformity constraints. 

We then generated 4,000 prompts  and produced high-resolution images with \texttt{DALL·E~3} using a robust, checkpointed pipeline (asynchronous batching, retries, and traceable file naming). The intrinsic safety moderation of \texttt{DALL·E~3} blocked a small subset of requests, primarily those containing dense named entities or historical and military references, which the system may have misinterpreted as potentially promoting violence, political propaganda, or disinformation.

\subsection{Evaluation Methods}
We conducted automatic and human evaluation using both automatically and expert-annotated accessibility, clarity, and style-related effects. Specifically, we used:  
\begin{itemize}[noitemsep, topsep=0pt, partopsep=0pt, parsep=0pt, left=0pt]
    \item \textbf{Inter-annotator agreement (IAA)}: Measured using Krippendorff's $\alpha$~\citep{krippendorff1970estimating} for human annotations;
    \item \textbf{Style identification}: Evaluated via Recall@3;
    \item \textbf{Human-computer correlation}: Measured using Pearson correlation coefficients; 
    \item \textbf{Composite accessibility scores}: Combining expert ratings into weighted indices of effectiveness.  
\end{itemize}
Statistical analysis included descriptive and comparative assessments across datasets, styles, and evaluation dimensions. 
Four expert annotators participated in this study, all with academic and professional expertise in accessibility and inclusive communication.  Their backgrounds involve specialization in barrier-free communication, Easy Language, audio description, subtitling, and text simplification.  Several have contributed to national research projects on accessible technologies and have practical experience designing and evaluating content for people with cognitive disabilities. All annotators signed informed consent forms and received detailed annotation guidelines. We compensated the expert annotators with a fair hour rate of 120 Swiss Francs.

The annotation study was originally planned with 4,000 generated images but reduced to 2,000 due to cost and workload constraints. The final set ensured each simplified sentence had images in all 10 styles, maintaining equal style representation. All images were renamed adopting a standardized numerical convention to ensure unbiased annotation, as original filenames contained style information that could influence decisions. A shared IAA set of 200 images (20 texts$\times$ 10 styles) was annotated by all four experts to assess agreement. The remaining 1,800 images were split evenly so each annotator received 200 IAA images plus 450 unique ones (650 images per annotator, 2,600 total annotations planned). In the end 976 annotations were completed by the four experts.

The annotation interface was implemented through a customized \texttt{Label Studio}\footnote{Apache-2.0 license, available at \url{https://github.com/HumanSignal/label-studio/}} configuration that presents the evaluation framework in a clear format optimized for expert assessment. The interface displays the simplified text at the top, followed by the image presented with zoom controls for detailed examination. The evaluation questions are organized into collapsible sections corresponding to distinct domains.  The configuration also ensured that annotation data was stored in structured formats suitable for subsequent statistical analysis. The interface included progress tracking which allowed the experts to track their progress and resume annotation across multiple sessions as needed. The complete set of evaluation questions is provided in Appendix~\ref{app:evaluation-questions} for reference.

\section{Results}
\label{sec:results}


A total of 976 annotations were collected from four experts (A, K, L, M), corresponding to 37.5\% of the initially planned 2,600 (Table~\ref{tab:annotation_completion_summary}). Despite lower coverage, the annotated sample provided sufficient data for meaningful analysis. Distribution across datasets was balanced (Table~\ref{tab:dataset_distribution_annotation}), ensuring findings are not skewed toward one text source.

\begin{table}[h]
    \centering
    \footnotesize
    \setlength{\tabcolsep}{6pt}
    \resizebox{0.8\columnwidth}{!}{
    \begin{tabular}{@{}lcc@{}}
        \toprule
        \textbf{Expert} & \textbf{\# Assigned} & \textbf{Completion Rate} \\
        \midrule
        Expert A & 650 & 38.5\% \\
        Expert K & 650 & 38.5\% \\
        Expert L & 650 & 30.8\% \\
        Expert M & 650 & 42.5\% \\
        \midrule
        \textbf{Total} & \textbf{2,600} & \textbf{37.5\%} \\
        \bottomrule
    \end{tabular}
    }
    
    \caption{Expert assignment and completion rates.}
    \label{tab:annotation_completion_summary}
\end{table}

\begin{table}[h]
    \centering
    \footnotesize
    \setlength{\tabcolsep}{6pt}
    \resizebox{0.8\columnwidth}{!}{
    \begin{tabular}{@{}lcc@{}}
        \toprule
        \textbf{Dataset} & \textbf{\# Annotation} & \textbf{Percentage} \\
        \midrule
        Wikipedia & 280 & 28.7\% \\
        SimPA     & 270 & 27.7\% \\
        ASSET     & 214 & 21.9\% \\
        OneStop   & 212 & 21.7\% \\
        \midrule
        \textbf{Total} & \textbf{976} & \textbf{100.0\%} \\
        \bottomrule
    \end{tabular}
    }
    
    \caption{Source distribution in completed dataset.}
    \label{tab:dataset_distribution_annotation}
\end{table}

\subsection{Scoring Overview}
Experts rated images across six dimensions with a maximum total of 100 points. Table~\ref{tab:dimension_contributions} summarizes mean scores and contributions. \textit{Ethics} dominated in expert evaluations (31.4\%), followed by \textit{Text Quality} (21.6\%). Visual dimensions such as \textit{Image Simplicity} and \textit{Text–Image Alignment} contributed less, reflecting both interpretive challenges and model limitations.

\begin{table}[h]
\centering
\small
\setlength{\tabcolsep}{3pt}
\resizebox{\columnwidth}{!}{
\begin{tabular}{@{}lccc@{}}
\toprule
\textbf{Dimension} & \textbf{Scale} & \textbf{Mean} & \textbf{Contribution} \\
\midrule
Image Simplicity       & 0--15  & 4.56  & 9.7\% \\
Image Quality          & 0--15  & 6.34  & 13.5\% \\
Text Simplicity        & 0--15  & 5.65  & 12.1\% \\
Text Quality           & 0--15  & 10.13 & 21.6\% \\
Ethics                 & 0--20  & 14.74 & 31.4\% \\
Text-Image Alignment   & 0--20  & 5.49  & 11.7\% \\
\bottomrule
\end{tabular}
}

\caption{Six evaluation dimensions used during expert annotation.}
\label{tab:dimension_contributions}
\end{table}

\begin{figure}[h]
    \centering
    \includegraphics[width=\columnwidth]{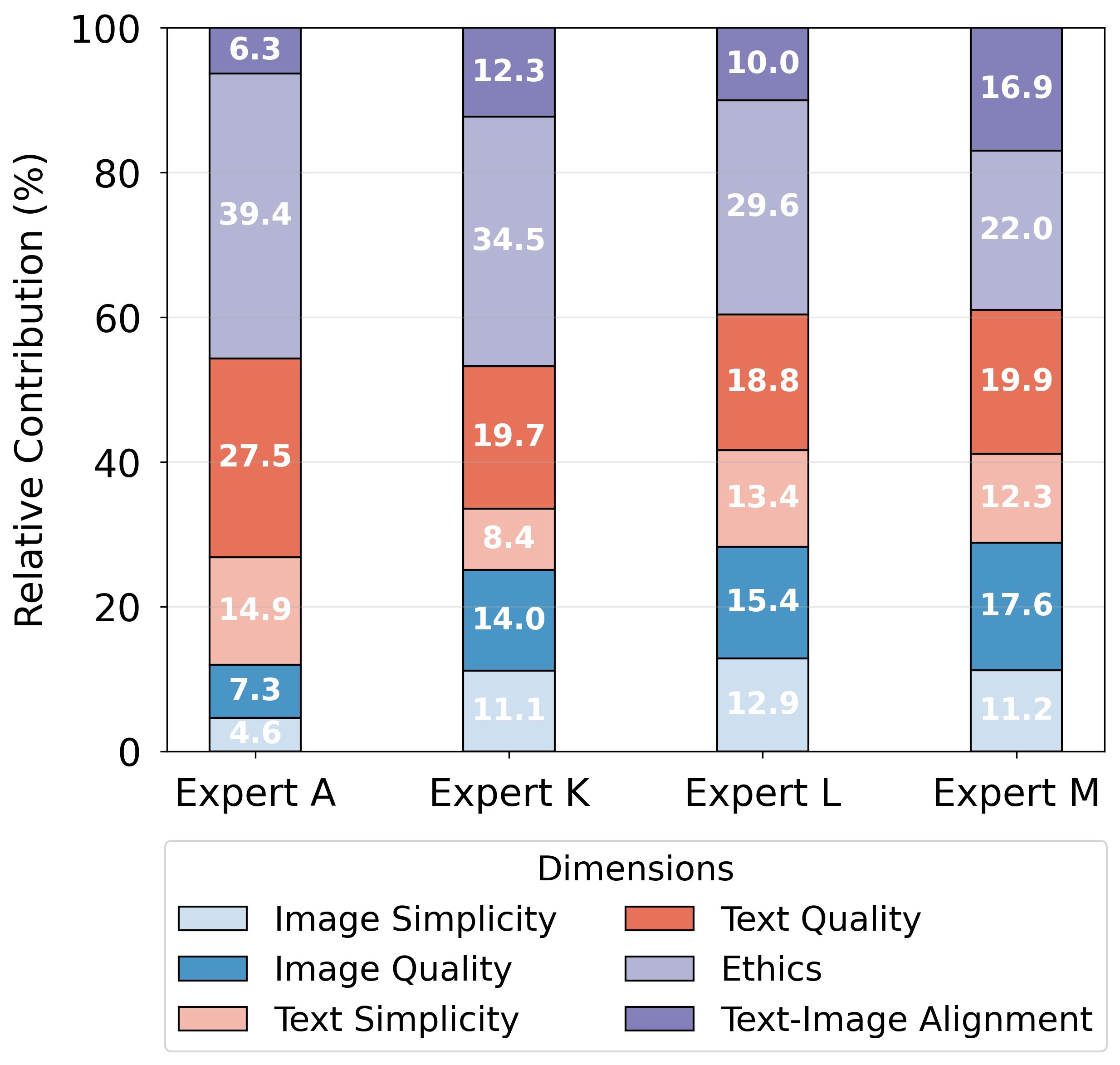}
    \caption{Relative contribution of each evaluation dimension per expert.}
    \label{fig:expert_examples}
\end{figure}

We observed that experts differed in evaluation style: Expert A was highly critical of visual dimensions but more lenient on \textit{Ethics}; Expert K was generous overall, with the highest mean scores; Expert L was most conservative, assigning the lowest averages; Expert M showed consistency, prioritizing \textit{Image Quality} and \textit{Text-Image Alignment}. 
Figure~\ref{fig:expert_examples} illustrates the composition of expert ratings.

\subsection{Inter-Annotator Agreement}
IAA was measured with Krippendorff’s Alpha across dimensions (Table~\ref{tab:iaa_overall}). \textit{Text Simplicity} achieved the highest agreement (up to $\alpha = 0.599$), suggesting a shared understanding of linguistic complexity. Agreement was weaker for \textit{Image Simplicity} and \textit{Text-Image Alignment}, and nearly absent for \textit{Image Quality} and \textit{Ethics}, reflecting subjective ratings.

\begin{table}[h]
\centering
\resizebox{\columnwidth}{!}{
\begin{tabular}{@{}lccc@{}}
\toprule
\textbf{Dimension} & \textbf{4 Experts} & \textbf{3+ Experts} & \textbf{2+ Experts} \\
\midrule
Text Simplicity       & 0.599 & 0.458 & 0.570 \\
Image Simplicity      & 0.486 & 0.374 & 0.282 \\
Text-Image Alignment  & 0.379 & 0.267 & 0.115 \\
Text Quality          & 0.086 & 0.228 & 0.223 \\
Ethics                & 0.018 & 0.053 & -0.214 \\
Image Quality         & 0.013 & -0.007 & 0.006 \\
\bottomrule
\end{tabular}
}

\caption{Krippendorff's Alpha across dimensions, reported for all and subgroup of experts.}
\label{tab:iaa_overall}
\end{table}

\subsection{Style Recognition Performance}
To study the influence of image styles for accessibility, we asked the experts to identify three image styles from the ten pre-defined style categories. Recall@3 averaged 47.3\%, with significant variance across experts (Table~\ref{tab:style_recall_experts}). Styles such as \textit{3D Rendered} and \textit{Retro} were easily recognized, while \textit{Artistic} and \textit{Technical} proved to be the most difficult ones (Table~\ref{tab:style_difficulty}).

\begin{table}[h]
\centering
\setlength{\tabcolsep}{6pt}
\resizebox{0.8\columnwidth}{!}{
\begin{tabular}{lccc}
\toprule
\textbf{Expert} & \textbf{\# Correct} & \textbf{\# Total} & \textbf{Recall@3} \\
\midrule
Expert A & 126 & 250 & 0.504 \\
Expert M & 139 & 276 & 0.504 \\
Expert K & 120 & 250 & 0.480 \\
Expert L & 77  & 200 & 0.385 \\
\midrule
\textbf{Average} & 462 & 976 & 0.473 \\
\bottomrule
\end{tabular}
}
\caption{Recall@3 for style recognition.}
\label{tab:style_recall_experts}
\end{table}

\begin{table}[h]
\centering
\resizebox{\columnwidth}{!}{
\begin{tabular}{lccc}
\toprule
\textbf{Style} & \textbf{\# Images} & \textbf{Recall@3} & \textbf{Tier} \\
\midrule
3D Rendered    & 103 & 81.6\% & Easy \\
Retro          & 100 & 76.0\% & Easy \\
Cartoon        & 109 & 69.7\% & Medium \\
Geometric      & 94  & 55.3\% & Medium \\
Realistic      & 86  & 44.2\% & Hard \\
Storybook      & 96  & 43.8\% & Hard \\
Digital Art    & 97  & 37.1\% & Hard \\
Minimalistic   & 88  & 34.1\% & Hard \\
Artistic       & 100 & 18.0\% & Very Hard \\
Technical      & 103 & 9.7\%  & Very Hard \\
\bottomrule
\end{tabular}
}
\caption{Style recognition difficulty comparison.}
\label{tab:style_difficulty}
\end{table}

\subsection{Human-Computer Correlation}

We compared CLIPScores with expert \textit{Text–Image Alignment} ratings. We observed weak but statistically significant Pearson correlation ($r=0.17$, $p<0.001$), improving after per-expert standardization (Table~\ref{tab:clip_correlation}). We observed that Expert A aligned most closely with CLIPScore ($r=0.251$). In addition, we found that \textit{Text-Image Alignment} varied by style, with strongest results for \textit{3D Rendered} and \textit{Artistic}, and weakest for \textit{Technical}.

\begin{table}[H]
\centering
\resizebox{0.8\columnwidth}{!}{
\begin{tabular}{lccc}
\toprule
\textbf{Analysis Level} & \textbf{$r$} & \textbf{$p$-value} & \textbf{Sig.} \\
\midrule
Overall (Raw)        & 0.133 & $<$0.001 & *** \\
Per-Expert Standard. & 0.173 & $<$0.001 & *** \\
Expert A             & 0.251 & $<$0.001 & *** \\
Expert M             & 0.152 & 0.011    & * \\
Expert L             & 0.146 & 0.040    & * \\
Expert K             & 0.133 & 0.036    & * \\
\bottomrule
\end{tabular}
}

\caption{Human-Computer correlation (Pearson $r$).}
\label{tab:clip_correlation}
\end{table}



\subsection{Expert-Specific Scoring Patterns}
\label{sec:expert-patterns}

\begin{figure}[htb]
  \centering
  \begin{minipage}[t]{0.48\linewidth}
    \centering
    \includegraphics[width=\linewidth]{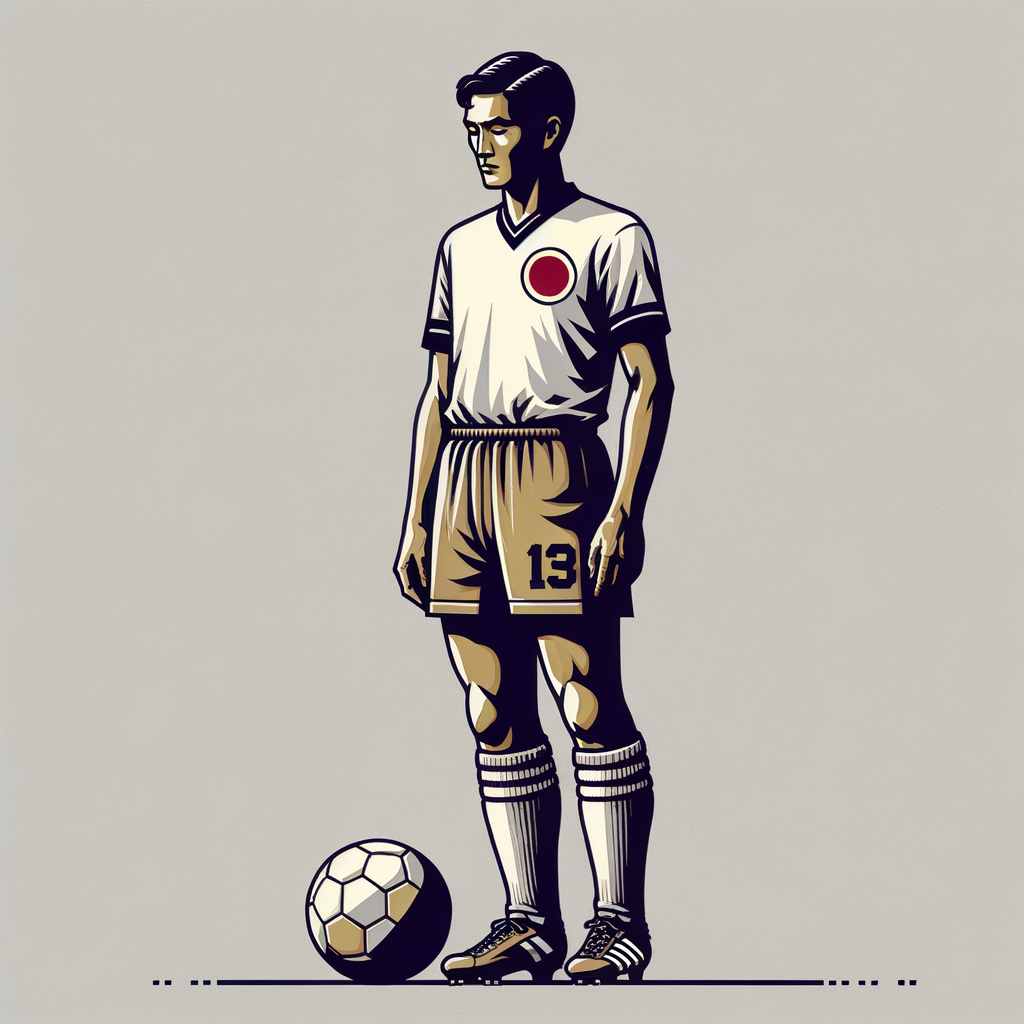}
    \vspace{0.25em}
    \footnotesize (a) \textbf{Expert A (Top)}
    
    \textit{``Kazuma Watanabe (born 10 August 1986) is a Japanese football player. He plays for Yokohama F. Marinos and Japan national team.''}
  \end{minipage}\hfill
  \begin{minipage}[t]{0.48\linewidth}
    \centering
    \includegraphics[width=\linewidth]{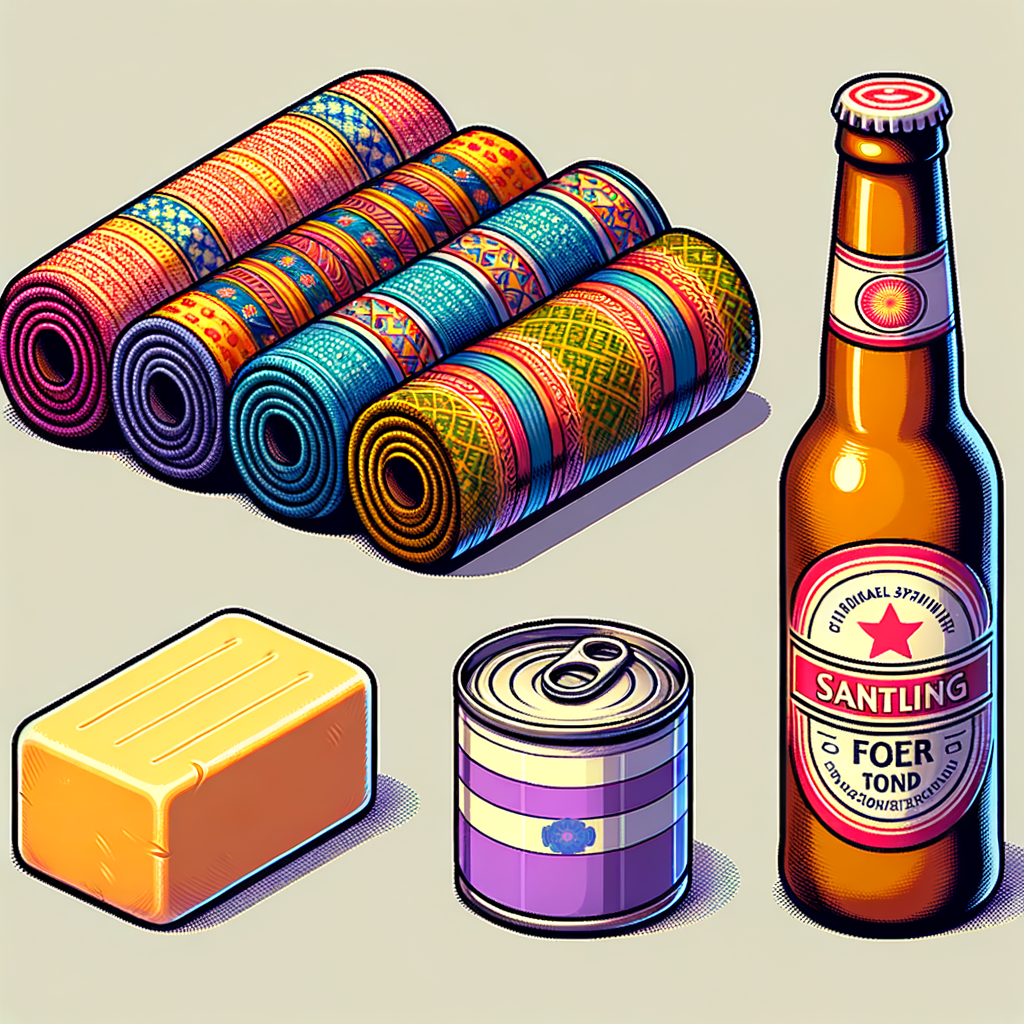}
    \vspace{0.25em}
    \footnotesize (b) \textbf{Expert K (Top)}
    
    \textit{``Bangui makes textiles, food products, beer, shoes, and soap.''}
  \end{minipage}

  \vspace{0.75em}

  \begin{minipage}[t]{0.48\linewidth}
    \centering
    \includegraphics[width=\linewidth]{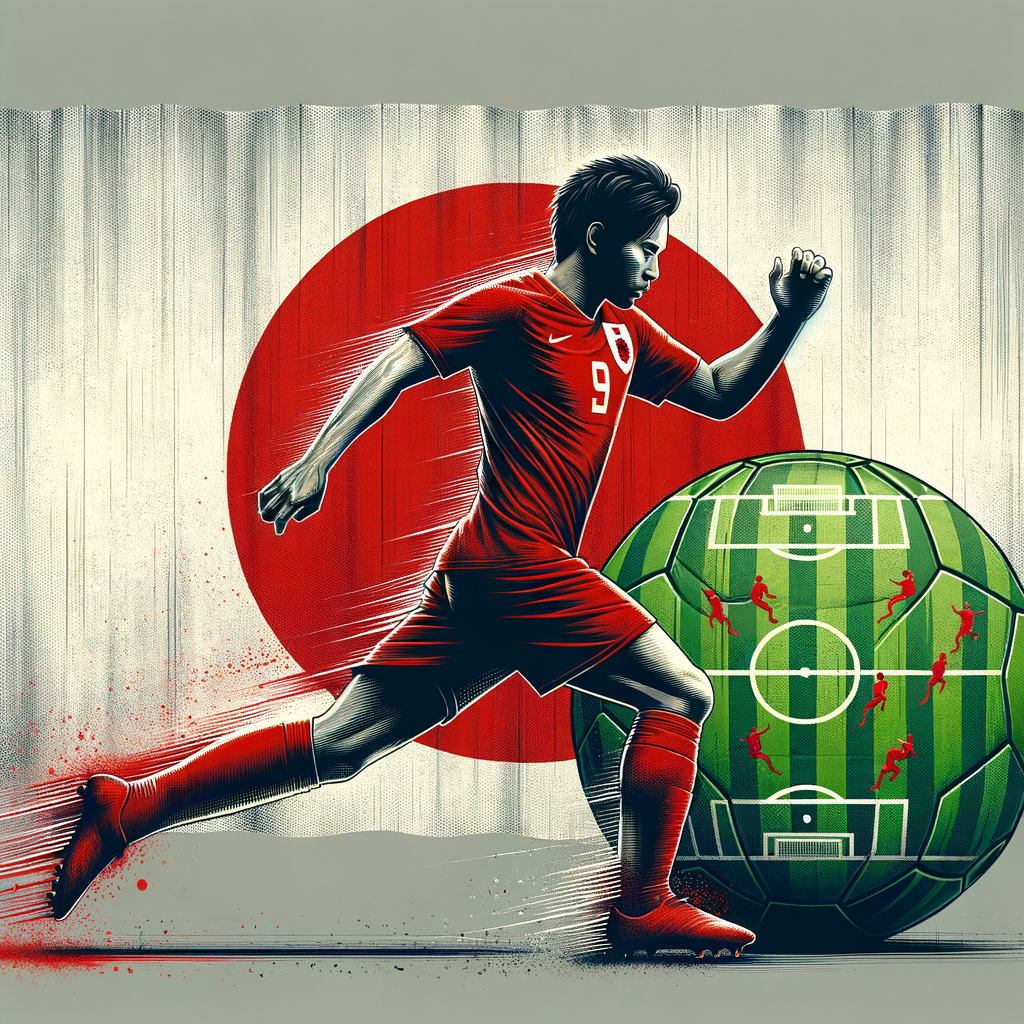}
    \vspace{0.25em}
    \footnotesize (c) \textbf{Expert L (Top)} 
    
    \textit{``Shunsuke Iwanuma (born 2 June 1988) is a Japanese football player. He plays for Consadole Sapporo.''}
  \end{minipage}\hfill
  \begin{minipage}[t]{0.48\linewidth}
    \centering
    \includegraphics[width=\linewidth]{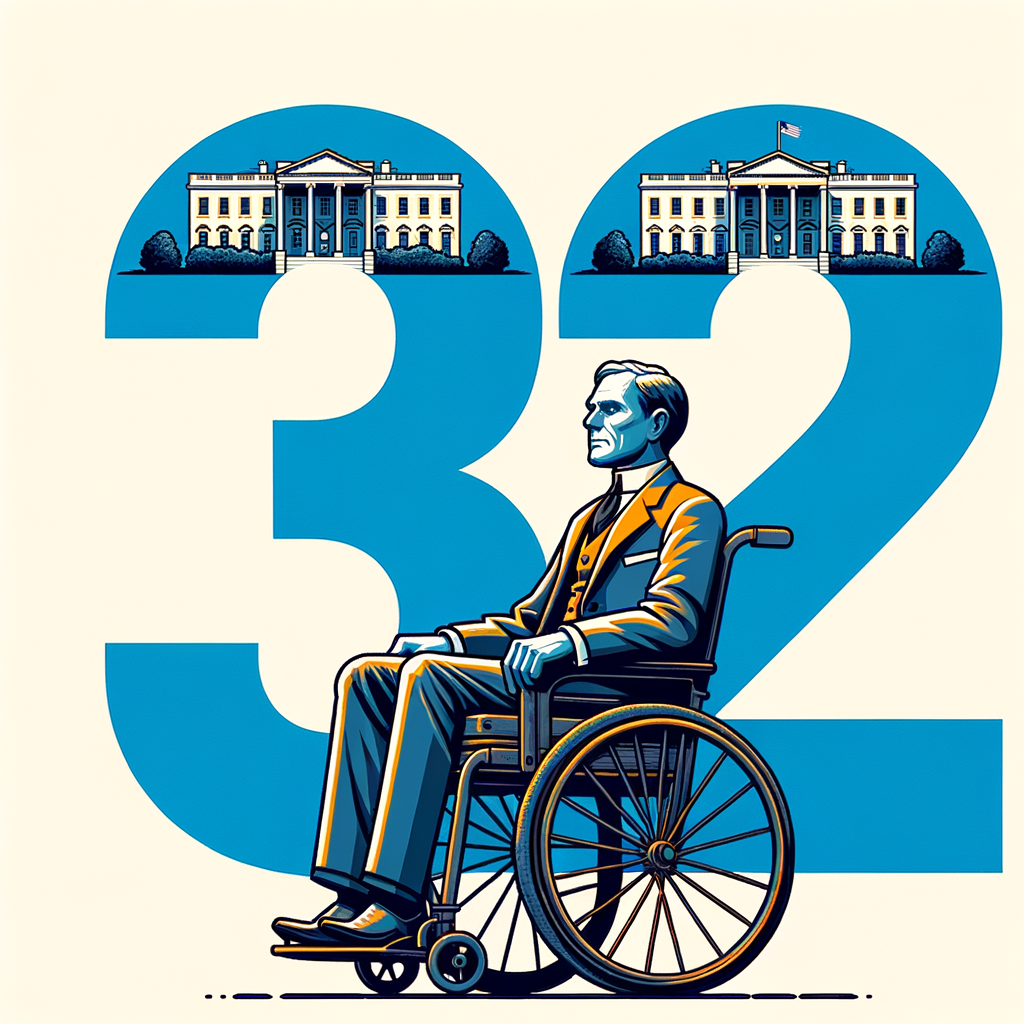}
    \vspace{0.25em}
    \footnotesize (d) \textbf{Expert M (Top)} 
    
    \textit{``Franklin Delano Roosevelt or FDR, was the 32nd President of the United States.''}
  \end{minipage}

  \caption{Highest-rated images by each expert, with corresponding simplified sentences.}
  \label{fig:expert_top_grid}
\end{figure}


\begin{figure}[htb]
  \centering
  \begin{minipage}[t]{0.48\linewidth}
    \centering
    \includegraphics[width=\linewidth]{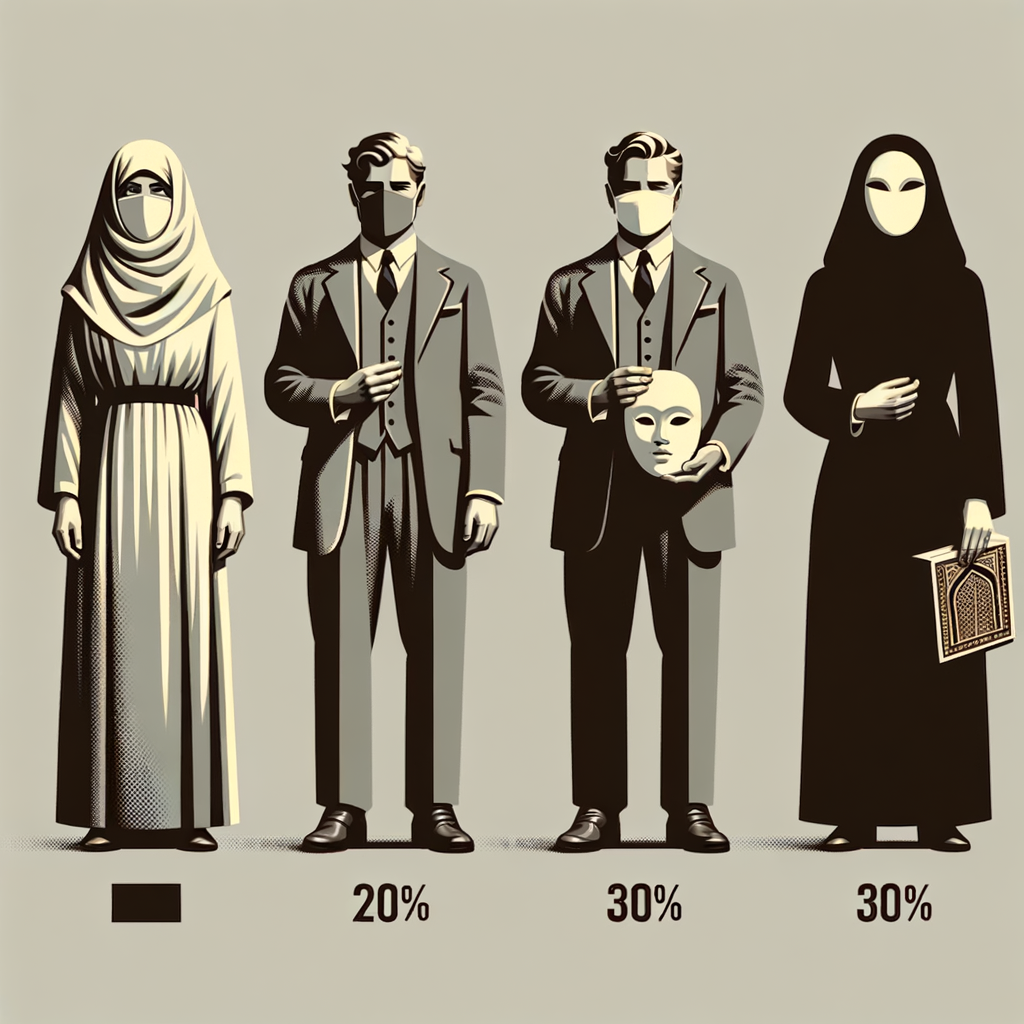}
    \vspace{0.25em}
    \footnotesize (a) \textbf{Expert A} (Bottom)
    
    \textit{``In Mask fetishism is persons wants to see another person wearing mask or taking off a mask.''}
  \end{minipage}\hfill
  \begin{minipage}[t]{0.48\linewidth}
    \centering
    \includegraphics[width=\linewidth]{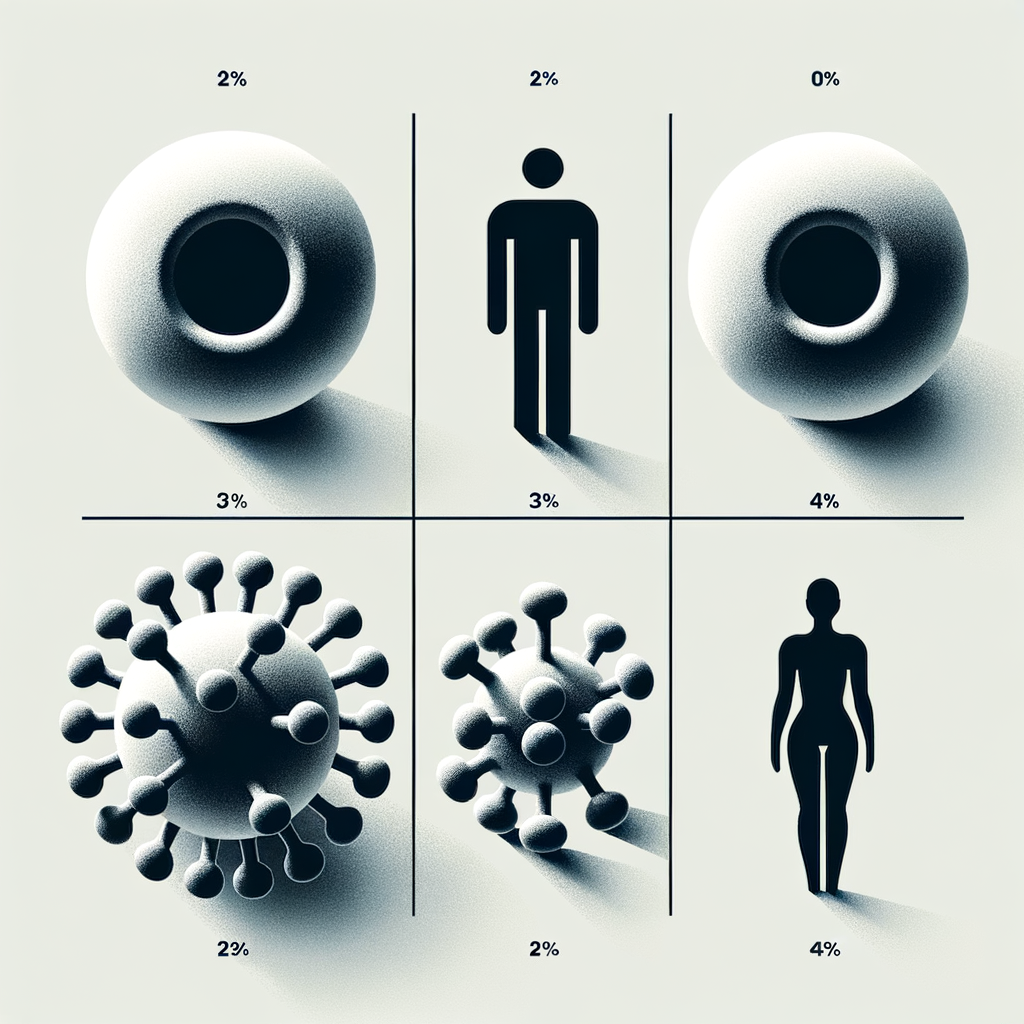}
    \vspace{0.25em}
    \footnotesize (b) \textbf{Expert K} (Bottom)
    
    \textit{``Cytomegalovirus (from the Greek cyto-, cell, and megalo-, large) is a viral genus of the Herpesviruses group in humans...''}
  \end{minipage}
  \caption{Lowest-rated images by Expert A and K.}
  \label{fig:expert_bottom_ak}
\end{figure}

\begin{figure}[htb]
  \centering
  \begin{minipage}[t]{0.48\linewidth}
    \centering
    \includegraphics[width=\linewidth]{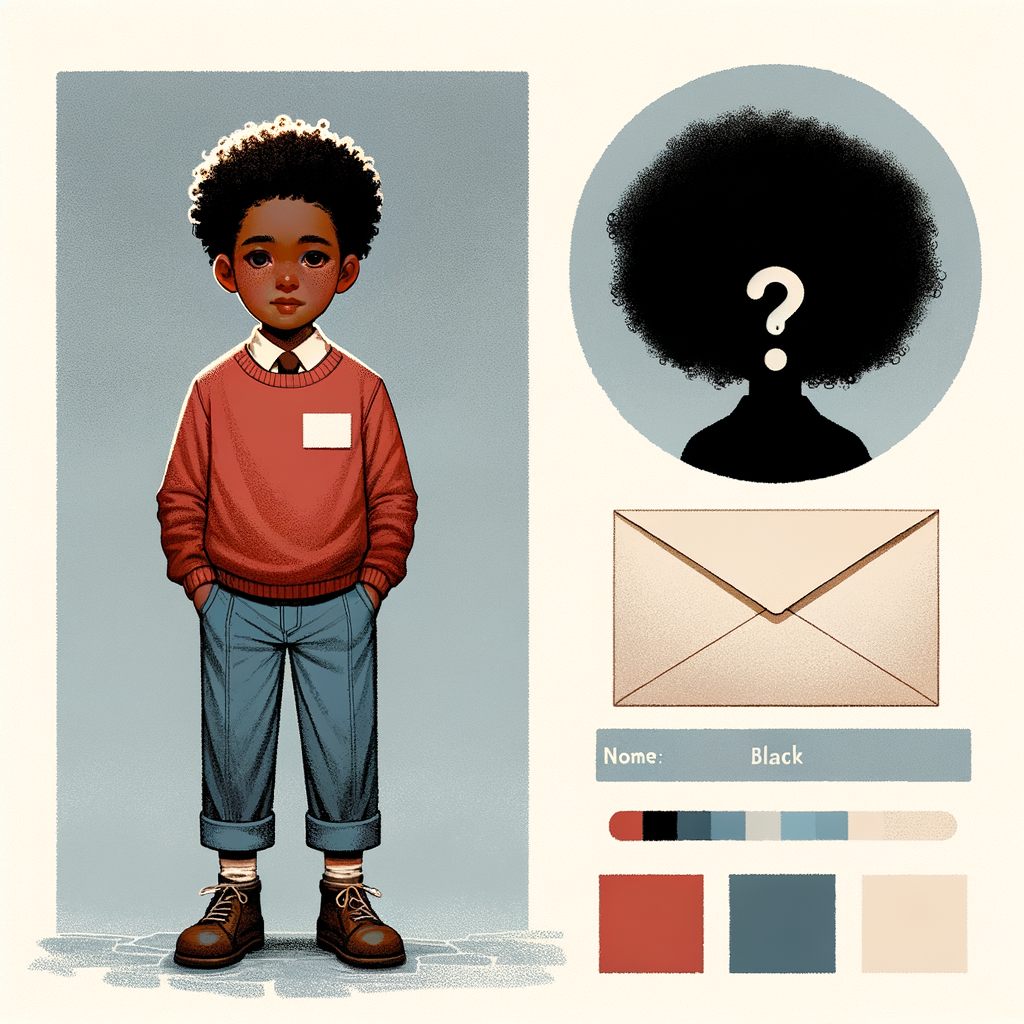}
    \vspace{0.25em}
    \footnotesize (c) \textbf{Expert L} (Bottom) 
    
    \textit{``One of the boyfriends wrote to me and said, Listen, she's not mad but Cynthia found out.''}
  \end{minipage}\hfill
  \begin{minipage}[t]{0.48\linewidth}
    \centering
    \includegraphics[width=\linewidth]{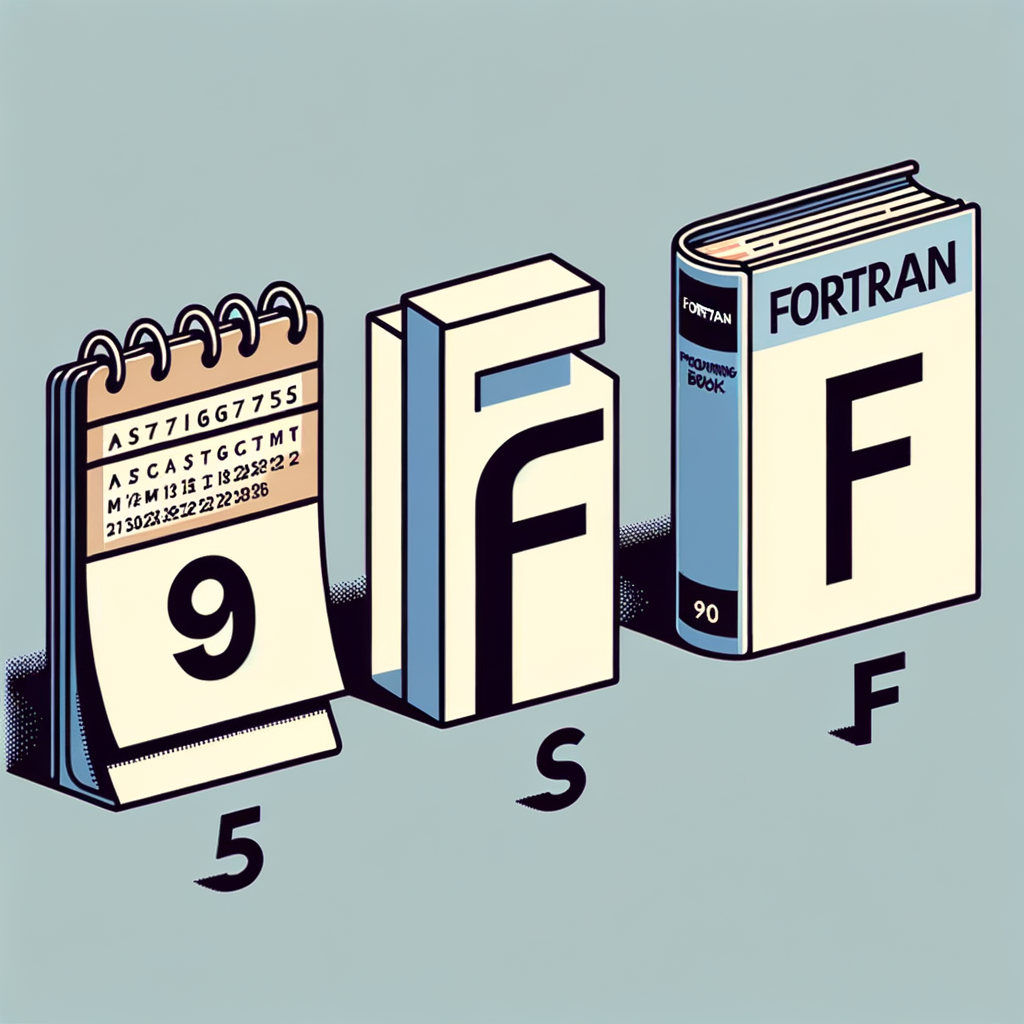}
    \vspace{0.25em}
    \footnotesize (d) \textbf{Expert M }(Bottom) 
    
    \textit{``Since Fortran 90, the capitalization has been abandoned. The published formal standards use Fortran.''}
  \end{minipage}
  \caption{Lowest-rated images by Expert L and M.}
  \label{fig:expert_bottom_lm}
\end{figure}

\begin{figure}[htb]
  \centering
  \begin{minipage}[t]{0.485\linewidth}
    \centering
    \includegraphics[width=\linewidth]{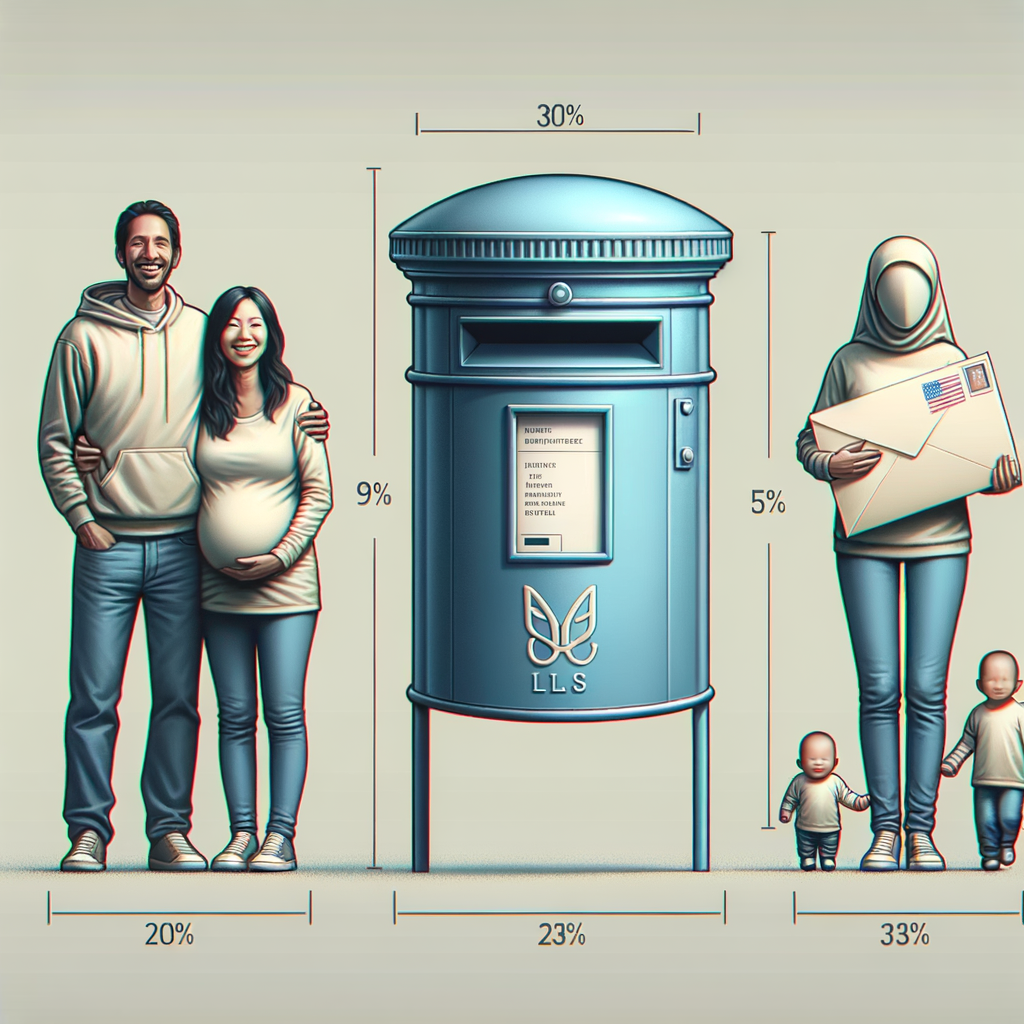}
    \vspace{0.25em}
    \footnotesize
    \textbf{Case 1 (0125)}\newline
    \emph{``Letterbox Service ...''}\newline
    Percentages not linked to text, cluttered layout, weak link between people and the concept. 
    Some experts gave very low Alignment/Quality, others still rewarded Ethics.
  \end{minipage}\hfill
  \begin{minipage}[t]{0.485\linewidth}
    \centering
    \includegraphics[width=\linewidth]{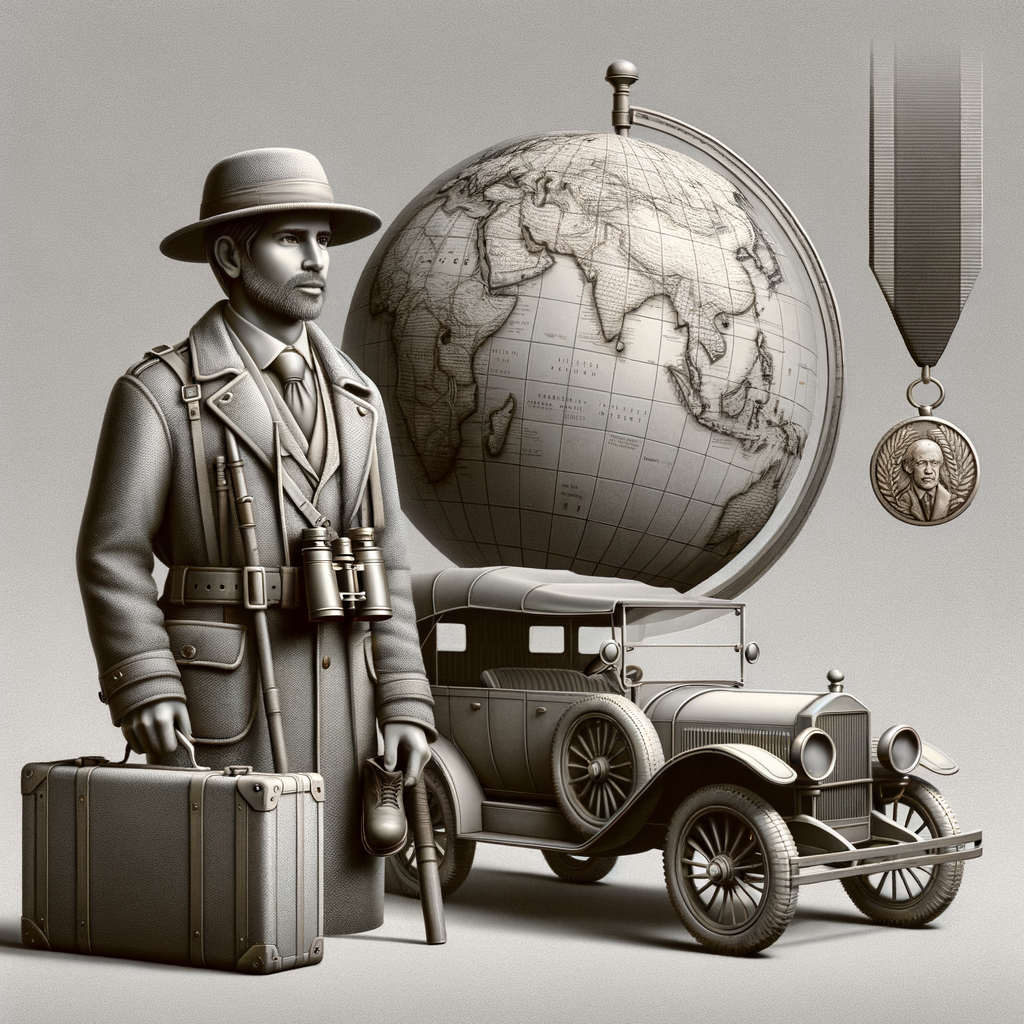}
    \vspace{0.25em}
    \footnotesize
    \textbf{Case 2 (0944)}\newline
    \emph{``He traveled over 200{,}000 miles ...''}\newline
    Vintage style with many props, missing legs, low contrast, and the number not shown. 
    Experts disagreed on Simplicity and Quality.
  \end{minipage}
  \vspace{-0.25em}
  \caption{Examples where experts strongly disagreed.}
  \label{fig:disagreement-cases}
\end{figure}

To further understand individual expert ratings, we conducted per-expert analysis on how they rate different evaluation dimensions, with a focus on how individual experts subjectively rate different evaluation dimensions.

We found that \textit{Image Quality} and \textit{Ethics} were the two dimensions with the lowest agreement. 
The main reason is that each expert used the scale differently. 
For example, Expert A was very strict with visuals but very generous in Ethics, 
while Expert M was the opposite. This systematic rating behavior explains why the same images received very different scores.

Figure~\ref{fig:expert_top_grid} to Figure~\ref{fig:disagreement-cases} present concrete examples of high and low scores given by experts. While some of these images might be potentially biased, these cases show that even with advanced VLMs, generating accessible visual interpretation for text simplification remains a challenging task, as it might further introduce confusion and misinterpretation of the text.

Building on these expert-annotated examples, we emphasize the importance of systematically examining the cultural and ethical biases present in VLMs when generating visuals for text simplification. Our findings suggest that even when accessibility constraints are technically enforced, subtle cultural cues, stereotypes, or contextual omissions can still emerge, potentially undermining the inclusiveness of the generated content. Future research should therefore not only focus on improving visual alignment and linguistic adequacy, but also on auditing and mitigating such biases through participatory evaluation and the inclusion of diverse user perspectives. Addressing these challenges is essential for developing truly inclusive, transparent, and socially responsible AI systems that can serve users with a broad range of linguistic and cognitive needs.

\FloatBarrier

\section{Discussion}
\label{sec:expert-disagreement}

\FloatBarrier


\subsection{Style and Dataset Impact for Accessibility}

To determine which styles and datasets best support accessibility, we combined expert ratings into composite accessibility scores. For visual styles, the score weighted \textit{Text–Image Alignment} (60\%), \textit{Image Simplicity} (25\%), and \textit{Image Quality} (15\%). For datasets, we focused on \textit{Text Quality} (50\%) and \textit{Text Simplicity} (50\%). Scores were scaled to a 0–100 range and averaged across experts.

\paragraph{Visual styles.} 
\textit{Retro} achieved the highest overall accessibility, performing strongly in both alignment and simplicity. \textit{Realistic} style also scored well, driven by high image quality and familiarity, while \textit{3D Rendered} provided the most consistent results. By contrast, abstract or heavily stylized categories such as \textit{Geometric}, \textit{Minimalistic}, and \textit{Artistic} ranked lowest, often due to clutter or ambiguous representation. These results suggest that concrete and familiar visuals are more effective for accessibility than abstract or decorative designs.

\paragraph{Data sources.} 
Wikipedia and ASSET ranked highest, with Wikipedia benefiting from factual clarity and wide coverage, and ASSET offering more consistency due to its accessibility-focused design. OneStopEnglish and SimPA performed less well, with lower simplicity scores and more complex phrasing. This indicates that general-purpose and collaboratively simplified corpora provide stronger material for accessible image generation.

The findings highlight that both style and dataset choice substantially affect accessibility outcomes. \textit{Retro} and \textit{Realistic} visuals, paired with simplified Wikipedia content, emerged as the most effective combination. Conversely, highly abstract styles or bureaucratic text sources may hinder accessibility, underscoring the need to carefully select both visual style and textual input when designing accessible multimodal content.

\subsection{Evaluation Agreement}

The evaluation highlights both strengths and weaknesses of template-based image generation for accessibility. The 100-point scoring framework showed that \textit{Ethics} and \textit{Text Quality} consistently received the highest ratings, together accounting for more than half of the total, likely because these dimensions had clearer definitions. By contrast, \textit{Image Simplicity} and \textit{Text–Image Alignment} contributed less and showed greater variability, pointing to interpretive difficulties and model limitations. Expert behavior differed noticeably: some applied stricter standards while others were more generous, underlining the need for normalization when aggregating scores.

Inter-annotator agreement further confirmed these differences. \textit{Text Simplicity} achieved the highest reliability, indicating a shared understanding of linguistic complexity. However, dimensions such as \textit{Image Quality} and \textit{Ethics} had very low or even negative agreement values, reflecting systematic disagreement rooted in subjective interpretations of what counts as ``quality'' or ``ethical''. These results suggest that clearer task guidelines, calibration, or refined criteria will be necessary to improve reproducibility in future studies.

Style recognition performance also highlighted challenges. Experts correctly identified the intended style in less than half of cases, which is reasonable given overlapping visual features across categories. Distinctive styles such as \textit{3D Rendered} or \textit{Retro} were easier to recognize, whereas \textit{Artistic} and \textit{Technical} were consistently difficult, revealing both model inconsistencies and unclear category boundaries. Style recognition therefore appears cognitively demanding and not fully reliable for accessibility-focused evaluations.




\section{Conclusion}

This work investigated template-based prompting for generating cognitively accessible images from simplified text. The \textit{Basic Object Focus} template proved most effective, showing that minimalism and object isolation enhance clarity. Wikipedia and ASSET emerged as the most suitable datasets, while \textit{Retro} and \textit{Realistic} visuals supported accessibility better than abstract styles. Expert evaluations highlighted systematic disagreement on subjective dimensions, and CLIPScores showed only weak alignment with human judgments. Overall, our findings demonstrate that structured prompting improves accessibility but current VLMs remain limited, underscoring the need for better models, clearer annotation protocols, and continued integration of human expertise.

\section*{Limitations}

Our study faced several constraints. Only 976 of the planned 2,000 annotations were collected, reducing statistical power and inter-annotator coverage. The expert panel was small (four annotators), making results sensitive to individual biases, and subjective dimensions such as \textit{Image Quality} and \textit{Ethics} showed strong disagreement. Despite prompt constraints, the model often produced cluttered layouts or text artifacts, and style recognition was hindered by ambiguous category boundaries. Finally, CLIPScore correlated only weakly with human judgments, raising concerns about its reliability for accessibility evaluation. These limitations highlight the need for clearer guidelines, improved models, and more diverse expert input in future work.

\section*{Ethical Statement}
All expert annotators involved in the human evaluation provided informed consent and received fair compensation for their participation. The evaluation framework included ethical safeguards designed to detect and flag potentially discriminatory or culturally insensitive content. Additionally, filtering mechanisms were applied during image generation to mitigate harmful or inappropriate outputs, and selected filtering results were documented to assess potential model biases.

The authors acknowledge that some example images presented in this work may still reflect unintended biases or misleading representations toward certain ethnic or cultural groups. These instances are discussed transparently to emphasize the importance of continuous bias assessment in developing inclusive AI systems.

\section*{Lay Summary}

Individuals with cognitive disabilities or reading difficulties often face challenges when processing complex visual and textual information. Current text-to-image models tend to prioritize photorealism or artistic creativity rather than accessibility, leaving a gap for users who require simplified and structured visuals.

Our study addresses this issue by integrating text simplification with structured prompting to generate cognitively accessible images. We designed five visual templates that control object number, spatial arrangement, and content complexity, while deliberately avoiding textual or abstract elements that may cause confusion.

This work contributes practical guidelines for producing accessible visual content. The proposed image styles and prompt templates can support policymakers, AI researchers, and assistive technology designers in creating visual materials that are both inclusive and cognitively accessible. By aligning text-to-image generation with accessibility principles, our approach ensures more equitable support for individuals with cognitive disabilities.

\section*{Acknowledgments}
This work was supported by the Swiss Innovation Agency Innosuisse, Flagship Inclusive Information and Communication Technology (IICT), funding no. PFFS-21-47. 
We also thank our experts Luisa Carrer, Martin Kapuss, Alexa Lintner at Zurich University of Applied Sciences (ZHAW), and Katrin Andermatt at traduko, for their valuable contributions to the study. Finally, we thank the anonymous reviewers for their constructive feedback on our work.

\bibliography{paper}

\begin{thebibliography}{25}
\providecommand{\natexlab}[1]{#1}

\bibitem[{Achiam et~al.(2023)Achiam, Adler, Agarwal, Ahmad, Akkaya, Aleman, Almeida, Altenschmidt, Altman, Anadkat et~al.}]{achiam2023gpt}
Josh Achiam, Steven Adler, Sandhini Agarwal, Lama Ahmad, Ilge Akkaya, Florencia~Leoni Aleman, Diogo Almeida, Janko Altenschmidt, Sam Altman, Shyamal Anadkat, and 1 others. 2023.
\newblock {GPT-4 Technical Report}.
\newblock \emph{arXiv preprint arXiv:2303.08774}.

\bibitem[{Alva-Manchego et~al.(2020)Alva-Manchego, Martin, Bordes, Scarton, Sagot, and Specia}]{alva2020asset}
Fernando Alva-Manchego, Louis Martin, Antoine Bordes, Carolina Scarton, Beno{\^\i}t Sagot, and Lucia Specia. 2020.
\newblock {ASSET: A Dataset for Tuning and Evaluation of Sentence Simplification Models with Multiple Rewriting Transformations}.
\newblock In \emph{Proceedings of the 58th Annual Meeting of the Association for Computational Linguistics}, pages 4668--4679.

\bibitem[{Ansch{\"u}tz et~al.(2024)Ansch{\"u}tz, Sylaj, and Groh}]{anschutz2024images}
Miriam Ansch{\"u}tz, Tringa Sylaj, and Georg Groh. 2024.
\newblock {Images Speak Volumes: User-Centric Assessment of Image Generation for Accessible Communication}.
\newblock In \emph{Proceedings of the Third Workshop on Text Simplification, Accessibility and Readability (TSAR 2024)}, pages 27--40.

\bibitem[{Betker et~al.(2023)Betker, Goh, Jing, Brooks, Wang, Li, Ouyang, Zhuang, Lee, Guo et~al.}]{betker2023improving}
James Betker, Gabriel Goh, Li~Jing, Tim Brooks, Jianfeng Wang, Linjie Li, Long Ouyang, Juntang Zhuang, Joyce Lee, Yufei Guo, and 1 others. 2023.
\newblock {Improving Image Generation with Better Captions}.
\newblock \emph{Computer Science. https://cdn. openai. com/papers/dall-e-3. pdf}, 2(3):8.

\bibitem[{Degraeuwe and Saggion(2022)}]{degraeuwe2022lexical}
Jasper Degraeuwe and Horacio Saggion. 2022.
\newblock {Lexical Simplification in Foreign Language Learning: Creating Pedagogically Suitable Simplified Example Sentences}.
\newblock In \emph{Proceedings of the Workshop on Text Simplification, Accessibility, and Readability (TSAR-2022)}, pages 98--110.

\bibitem[{Espinosa-Zaragoza et~al.(2023)Espinosa-Zaragoza, Abreu-Salas, Moreda, and Palomar}]{espinosa2023automatic}
Isabel Espinosa-Zaragoza, Jos{\'e} Abreu-Salas, Paloma Moreda, and Manuel Palomar. 2023.
\newblock {Automatic Text Simplification for People with Cognitive Disabilities: Resource Creation within the {C}lear{T}ext Project}.
\newblock In \emph{Proceedings of the Second Workshop on Text Simplification, Accessibility and Readability}, pages 68--77.

\bibitem[{Gao et~al.(2025)Gao, Johnson, Froehlich, Carrer, and Ebling}]{gao2025evaluting}
Yingqiang Gao, Kaede Johnson, David Froehlich, Luisa Carrer, and Sarah Ebling. 2025.
\newblock {Evaluating the Effectiveness of Direct Preference Optimization for Personalizing German Automatic Text Simplifications for Persons with Intellectual Disabilities}.
\newblock \emph{arXiv preprint arXiv:2507.01479}.

\bibitem[{Geislinger et~al.(2023)Geislinger, Pourasad, G{\"u}l, Djahangir, Yimam, Remus, and Biemann}]{geislinger2023multi}
Robert Geislinger, Ali~Ebrahimi Pourasad, Deniz G{\"u}l, Daniel Djahangir, Seid~Muhie Yimam, Steffen Remus, and Chris Biemann. 2023.
\newblock {Multi-Modal Learning Application - Support Language Learners with NLP Techniques and Eye-Tracking}.
\newblock In \emph{Proceedings of the 1st Workshop on Linguistic Insights from and for Multimodal Language Processing}, pages 6--11.

\bibitem[{Hessel et~al.(2021)Hessel, Holtzman, Forbes, Bras, and Choi}]{hessel2021clipscore}
Jack Hessel, Ari Holtzman, Maxwell Forbes, Ronan~Le Bras, and Yejin Choi. 2021.
\newblock {CLIPScore: A Reference-free Evaluation Metric for Image Captioning}.
\newblock \emph{arXiv preprint arXiv:2104.08718}.

\bibitem[{Heusel et~al.(2017)Heusel, Ramsauer, Unterthiner, Nessler, and Hochreiter}]{heusel2017gans}
Martin Heusel, Hubert Ramsauer, Thomas Unterthiner, Bernhard Nessler, and Sepp Hochreiter. 2017.
\newblock {GANs Trained by a Two Time-Scale Update Rule Converge to a Local Nash Equilibrium}.
\newblock \emph{Advances in neural information processing systems}, 30.

\bibitem[{Hu et~al.(2023)Hu, Liu, Kasai, Wang, Ostendorf, Krishna, and Smith}]{hu2023tifa}
Yushi Hu, Benlin Liu, Jungo Kasai, Yizhong Wang, Mari Ostendorf, Ranjay Krishna, and Noah~A Smith. 2023.
\newblock {TIFA: Accurate and Interpretable Text-to-Image Faithfulness Evaluation with Question Answering}.
\newblock In \emph{Proceedings of the IEEE/CVF International Conference on Computer Vision}, pages 20406--20417.

\bibitem[{Krippendorff(1970)}]{krippendorff1970estimating}
Klaus Krippendorff. 1970.
\newblock {Estimating the Reliability, Systematic Error and Random Error of Interval Data}.
\newblock \emph{Educational and Psychological Measurement}, 30(1):61--70.

\bibitem[{Li et~al.(2025)Li, Arase, and Crespi}]{li2025aligning}
Guanlin Li, Yuki Arase, and Noel Crespi. 2025.
\newblock {Aligning Sentence Simplification with {ESL} Learner{'}s Proficiency for Language Acquisition}.
\newblock In \emph{Proceedings of the 2025 Conference of the Nations of the Americas Chapter of the Association for Computational Linguistics: Human Language Technologies (Volume 1: Long Papers)}, pages 492--507.

\bibitem[{Lin et~al.(2014)Lin, Maire, Belongie, Hays, Perona, Ramanan, Doll{\'a}r, and Zitnick}]{lin2014microsoft}
Tsung-Yi Lin, Michael Maire, Serge Belongie, James Hays, Pietro Perona, Deva Ramanan, Piotr Doll{\'a}r, and C~Lawrence Zitnick. 2014.
\newblock {Microsoft COCO: Common Objects in Context}.
\newblock In \emph{Computer vision--ECCV 2014: 13th European conference, zurich, Switzerland, September 6-12, 2014, proceedings, part v 13}, pages 740--755. Springer.

\bibitem[{Madina et~al.(2023)Madina, Gonzalez-Dios, and Siegel}]{madina2023easy}
Margot Madina, Itziar Gonzalez-Dios, and Melanie Siegel. 2023.
\newblock {Easy-to-Read in Germany: A Survey on its Current State and Available Resources}.
\newblock \emph{arXiv preprint arXiv:2306.03189}.

\bibitem[{Marturi and Elwazzan(2025)}]{marturi2025llm}
Krishna~Chaitanya Marturi and Heba Elwazzan. 2025.
\newblock {LLM-Guided Planning and Summary-Based Scientific Text Simplification: DS@GT at CLEF 2025 SimpleText}.
\newblock \emph{arXiv preprint arXiv:2508.11816}.

\bibitem[{Plummer et~al.(2015)Plummer, Wang, Cervantes, Caicedo, Hockenmaier, and Lazebnik}]{plummer2015flickr30k}
Bryan~A Plummer, Liwei Wang, Chris~M Cervantes, Juan~C Caicedo, Julia Hockenmaier, and Svetlana Lazebnik. 2015.
\newblock {Flickr30k Entities: Collecting Region-to-Phrase Correspondences for Richer Image-to-Sentence Models}.
\newblock In \emph{Proceedings of the IEEE international conference on computer vision}, pages 2641--2649.

\bibitem[{Radford et~al.(2021)Radford, Kim, Hallacy, Ramesh, Goh, Agarwal, Sastry, Askell, Mishkin, Clark et~al.}]{radford2021learning}
Alec Radford, Jong~Wook Kim, Chris Hallacy, Aditya Ramesh, Gabriel Goh, Sandhini Agarwal, Girish Sastry, Amanda Askell, Pamela Mishkin, Jack Clark, and 1 others. 2021.
\newblock {Learning Transferable Visual Models from Natural Language Supervision}.
\newblock In \emph{International conference on machine learning}, pages 8748--8763. PmLR.

\bibitem[{Scarton et~al.(2018)Scarton, Paetzold, and Specia}]{scarton2018simpa}
Carolina Scarton, Gustavo Paetzold, and Lucia Specia. 2018.
\newblock {Simpa: A Sentence-Level Simplification Corpus for the Public Administration Domain}.
\newblock In \emph{Proceedings of the Eleventh International Conference on Language Resources and Evaluation (LREC 2018)}.

\bibitem[{Singh et~al.(2023)Singh, Zouhar, and Sachan}]{singh2023enhancing}
Janvijay Singh, Vil{\'e}m Zouhar, and Mrinmaya Sachan. 2023.
\newblock {Enhancing Textbooks with Visuals from the Web for Improved Learning}.
\newblock \emph{arXiv preprint arXiv:2304.08931}.

\bibitem[{Sun et~al.(2020)Sun, Lin, and Wan}]{sun2020helpfulness}
Renliang Sun, Zhe Lin, and Xiaojun Wan. 2020.
\newblock {On The Helpfulness of Document Context To Sentence Simplification}.
\newblock In \emph{Proceedings of the 28th International Conference on Computational Linguistics}, pages 1411--1423.

\bibitem[{Vajjala and Lu{\v{c}}i{\'c}(2018)}]{vajjala2018onestopenglish}
Sowmya Vajjala and Ivana Lu{\v{c}}i{\'c}. 2018.
\newblock {OneStopEnglish Corpus: A New Corpus for Automatic Readability Assessment and Text Simplification}.
\newblock In \emph{Proceedings of the thirteenth workshop on innovative use of NLP for building educational applications}, pages 297--304.

\bibitem[{Wang et~al.(2022)Wang, Schneider, Alacam, Chaudhury, and Biemann}]{wang2022motif}
Xintong Wang, Florian Schneider, {\"O}zge Alacam, Prateek Chaudhury, and Chris Biemann. 2022.
\newblock {MOTIF: Contextualized Images for Complex Words to Improve Human Reading}.
\newblock In \emph{Proceedings of the Thirteenth Language Resources and Evaluation Conference}, pages 2468--2477.

\bibitem[{Yawiloeng(2022)}]{yawiloeng2022using}
Rattana Yawiloeng. 2022.
\newblock {Using Instructional Scaffolding and Multimodal Texts to Enhance Reading Comprehension: Perceptions and Attitudes of EFL Students.}
\newblock \emph{Journal of Language and Linguistic Studies}, 18(2):877--894.

\bibitem[{Zhong et~al.(2020)Zhong, Jiang, Xu, and Li}]{zhong2020discourse}
Yang Zhong, Chao Jiang, Wei Xu, and Junyi~Jessy Li. 2020.
\newblock {Discourse Level Factors for Sentence Deletion in Text Simplification}.
\newblock In \emph{Proceedings of the AAAI conference on artificial intelligence}, volume~34, pages 9709--9716.

\end{thebibliography}

\newpage
\onecolumn
\appendix

\section{Overview of Dataset Sources}
\label{app:expert-examples}




    

\setcounter{table}{0}
\renewcommand{\thetable}{A\arabic{table}}
\begin{table}[htbp]
    \centering
    \footnotesize
    \setlength{\tabcolsep}{4pt}
    \resizebox{\columnwidth}{!}{
    \begin{tabular}{@{}llp{6.2cm}ll@{}}
        \toprule
        \textbf{Dataset} & \textbf{Domain} & \textbf{Size} & \textbf{Year} & \textbf{Level} \\
        \midrule
        \multirow{1}{*}{ASSET} & \multirow{1}{*}{Wikipedia} & 23,590 simplifications for 2,359 original sentences & \multirow{1}{*}{2020} & \multirow{1}{*}{Sentence} \\

        OneStopEnglish & News & 189 articles (567 texts) at three reading levels & 2018 & Document \& Sentence \\

        \multirow{2}{*}{SimPA} & \multirow{2}{*}{Web} & 1,100 sentences with 3 lexical and 1 syntactic simplification each & \multirow{2}{*}{2018} & \multirow{2}{*}{Sentence} \\

        Wikipedia (w/o context) & Wikipedia & 110K (with context) and 41K (without context) & 2020 & Sentence \\
        \bottomrule
    \end{tabular}
    }
    \caption{Overview of Text Simplification Datasets.}
    \label{tab:ch3_datasets}
\end{table}

\section{Prompt Templates used for Image Generation}
\label{app:prompt-templates}

This appendix provides the complete set of prompt templates developed for structured text-to-image generation. Each template implements specific accessibility constraints while maintaining semantic alignment with simplified text inputs.

\subsection{Basic Object Focus Template}
\begin{tcolorbox}[title=Basic Object Focus - Specific Prompt Instructions, colback=gray!5, colframe=gray!60!black, breakable]
\begin{itemize}[label=\textbullet]
  \item Do not align or group objects (arrange them with neutral positioning).
  \item Avoid any suggestion of scene, narrative, or sequence.
  \item Ensure all objects are visually equal.
  \item No object should stand out more than the others.
  \item Background must be uniform and simple (e.g., white or gray).
  \item Emphasize maximum spacing between all objects.
\end{itemize}
\end{tcolorbox}

\subsection{Contextual Scene Template}
\begin{tcolorbox}[title=Contextual Scene - Specific Prompt Instructions, colback=gray!5, colframe=gray!60!black, breakable]
\begin{itemize}[label=\textbullet]
  \item Arrange all objects in a straight, horizontal line.
  \item Use a single perspective, no variation in object size or depth.
  \item Maintain equal size across all objects to avoid depth illusion.
  \item Include one minimal environmental element (e.g., surface, wall) when needed.
  \item Keep at least 20\% spacing between each object to preserve separation.
\end{itemize}
\end{tcolorbox}

\subsection{Educational Layout Template}
\begin{tcolorbox}[title=Educational Layout - Specific Prompt Instructions, colback=gray!5, colframe=gray!60!black, breakable]
\begin{itemize}[label=\textbullet]
  \item Arrange objects in a strict left-to-right horizontal sequence.
  \item Visually connect each object to the next with a line or arrow.
  \item Gradually reduce object size from left to right by 10–15\%.
  \item Include a visible numeric marker (1, 2, 3...) near each object.
  \item Limit the maximum object count to 4 to maintain consistency.
  \item Narrow spacing slightly with each subsequent object to guide visual flow.
\end{itemize}
\end{tcolorbox}

\subsection{Multi-Level Detail Template}
\begin{tcolorbox}[title=Multi-Level Detail - Specific Prompt Instructions, colback=gray!5, colframe=gray!60!black, breakable]
\begin{itemize}[label=\textbullet]
  \item Place objects across exactly three spatial layers: foreground, midground, and background.
  \item Foreground objects must be 2× larger than midground objects.
  \item Midground objects must be 2× larger than background objects.
  \item Each layer must use a unique lightness or brightness level.
  \item Position layers vertically: foreground at the bottom, background at the top.
  \item Avoid horizontal alignment across layers to emphasize separation.
\end{itemize}
\end{tcolorbox}

\subsection{Grid Layout Template}
\begin{tcolorbox}[title=Grid Layout - Specific Prompt Instructions, colback=gray!5, colframe=gray!60!black, breakable]
\begin{itemize}[label=\textbullet]
  \item Choose a 2×2 or 3×3 grid structure, depending on object count.
  \item Place one object per cell, centered precisely.
  \item Use equal-sized cells with clearly defined, thick borders.
  \item Ensure all objects are the same size and prominence.
  \item Maintain at least 25\% margin around each object within its cell.
  \item Do not allow diagonal, overlapping, or asymmetrical arrangements.
\end{itemize}
\end{tcolorbox}

\subsection{General Accessibility Constraints}
All templates shared the following baseline requirements to ensure cognitive accessibility:
\begin{itemize}
\item \textbf{Object count control}: 3--5 distinct objects per image to avoid cognitive overload
\item \textbf{Spatial separation}: Minimum spacing requirements to enhance visual clarity
\item \textbf{Content restrictions}: Exclusion of text, numbers, abstract elements, or cultural bias
\item \textbf{Background simplicity}: Plain or neutral backgrounds to minimize distraction
\item \textbf{Visual equality}: Balanced prominence across objects unless explicitly specified
\end{itemize}

\clearpage

\section{Example JSON Entry}
\subsection{Example from the Dataset}
\label{app:datasets}
This appendix shows the structure of our compiled text-to-image TS corpus. Each entry contains the original and simplified sentence pairs along with metadata for traceability and analysis. This is an example from the final dataset of the 96th entry sampled from the ASSET corpus:

\begin{tcolorbox}[
    title=Dataset Structure Example,
    colback=gray!3,
    colframe=gray!60!black,
    breakable,
    enhanced,
    sharp corners,
    fonttitle=\bfseries\large,
    left=6pt,
    right=6pt,
    top=10pt,
    bottom=10pt,
    arc=2pt,
    boxsep=3pt
]

\vspace{1em} 
\noindent
\begin{minipage}{\textwidth}
\begin{lstlisting}[ 
    escapeinside={(*}{*)},
    basicstyle=\footnotesize\ttfamily,
    breaklines=true,
    frame=none,
    backgroundcolor=\color{white},
    xleftmargin=0pt,
    xrightmargin=0pt
]
{ 
  "id": "(*\textcolor{black}{asset\_069}*)",
  "(*\textcolor[HTML]{0000B2}{\textbf{dataset}}*)": "ASSET",
  "(*\textcolor[HTML]{0000B2}{\textbf{domain}}*)": "Wikipedia",
  "(*\textcolor[HTML]{0000B2}{\textbf{original}}*)": "The Odyssey is an ancient Greek epic poem attributed to Homer.",
  "(*\textcolor[HTML]{0000B2}{\textbf{simplified}}*)": "The Odyssey is an old Greek poem about Homer.",
  "(*\textcolor[HTML]{0000B2}{\textbf{length\_original}}*)": (*\textcolor{black}{12}*),
  "(*\textcolor[HTML]{0000B2}{\textbf{length\_simplified}}*)": (*\textcolor{black}{10}*)
}
\end{lstlisting}
\end{minipage}
\vspace{1em} 

\end{tcolorbox}

\subsection{Multiple Style Prompts for Single Sentence}

\label{app:prompt-example}
This appendix illustrates how our template-based framework generates style-specific prompts from a single simplified sentence. The example demonstrates the systematic application of accessibility constraints across ten distinct visual styles while maintaining semantic consistency.

\begin{tcolorbox}[
    title=Multiple Style Prompts for Single Sentence,
    colback=gray!3,
    colframe=gray!60!black,
    breakable,
    enhanced,
    sharp corners,
    fonttitle=\bfseries\large,
    left=6pt,
    right=6pt,
    top=10pt,
    bottom=10pt,
    arc=2pt,
    boxsep=3pt
]

\begin{lstlisting}[
    escapeinside={(*}{*)},
    basicstyle=\footnotesize\ttfamily,
    breaklines=true,
    frame=none,
    backgroundcolor=\color{white},
    xleftmargin=0pt,
    xrightmargin=0pt
]
{
  "index": (*\textcolor{black}{71}*),
  "id": "wikipedia_(*\textcolor{black}{387}*)", 
  "simplified_text": "Originally, a pie made with any kind of meat and mashed potato was called a cottage pie.",
  "dataset_source": "Wikipedia",
  "template_prompts": [
    {
      "style": "Cartoon",
      "prompt": "Generate a cartoon-(*\textcolor{black}{style}*) image with a light gray background. Include four distinct objects: A whole, uncooked piece of meat (such as a steak or a chicken drumstick). A knife and a fork, indicating the meat is ready to be cut and cooked. A bowl of raw, unpeeled potatoes. Each object should be scaled to similar sizes with no more than (*\textcolor{black}{10}*)% variation..."
    },
    {
      "style": "Realistic",
      "prompt": "Create a realistic image with a light gray background, showcasing four distinct objects: A piece of raw meat (like a steak or chicken breast) symbolizing 'any kind of meat'. A fresh, whole potato. A bowl of mashed potatoes. A traditional cottage or small house..."
    },
    {
      "style": "Artistic", 
      "prompt": "Produce an image in an Artistic (*\textcolor{black}{style}*) featuring the following elements in a clear and simple layout against a light-gray background: A classic pie dish, A piece of meat, A potato, A small cottage. Arrange with (*\textcolor{black}{30}*)% spacing between objects..."
    },
    {
      "style": "Minimalistic",
      "prompt": "Generate a minimalistic image with four objects: an empty pie dish, uncooked minced meat, ready-to-eat mashed potato, and a simple cottage representation. The pie dish should be empty, the minced meat should be uncooked and the mashed potato should look ready to eat. Each object should be scaled to similar sizes with no more than (*\textcolor{black}{10}*)% variation..."
    },
    {
      "style": "Digital Art",
      "prompt": "Create a digital art image consisting of four distinct objects: a pie, a piece of meat, a potato, and a simple cottage. Arrange these objects on a light gray background with a minimum of (*\textcolor{black}{30}*)% spacing between them. Maintain consistent sizing within 10% variation..."
    },
    {
      "style": "(*\textcolor{black}{3}*)D Rendered", 
      "prompt": "Generate an image in a (*\textcolor{black}{3}*)D rendered (*\textcolor{black}{style}*) on a neutral light-grey background. The image should contain four distinct objects: a pie dish, raw meat, a potato, and a cottage. Apply (*\textcolor{black}{30}*)% minimum spacing and maintain size consistency within (*\textcolor{black}{10}*)% variation..."
    },
    {
      "style": "Geometric",
      "prompt": "Generate a geometric (*\textcolor{black}{style}*) image composed of four distinct objects: a circular pie, a slice of pie, a piece of meat, and a mashed potato. Use simple geometric shapes and maintain (*\textcolor{black}{30}*)% spacing between elements..."
    },
    {
      "style": "Retro",
      "prompt": "Generate an image in a Retro (*\textcolor{black}{style}*) that depicts the following scene: A whole pie with a distinguishable crust on a light gray background. The pie should be depicted in a simple, stylized way with clear boundaries and vintage aesthetic..."
    },
    {
      "style": "Storybook", 
      "prompt": "Generate an image in a Storybook (*\textcolor{black}{style}*) on a solid light grey background. The picture should include four distinct objects: a whole uncooked pie, a separate piece of uncooked meat, a pile of mashed potatoes, and a small cottage. Each object should be scaled to a similar size with a (*\textcolor{black}{10}*)% variation allowance..."
    },
    {
      "style": "Technical",
      "prompt": "Create an image on a light gray background. The image should contain four distinct objects: a pie dish, a piece of meat, a potato, and a mashed potato. Use technical illustration (*\textcolor{black}{style}*) with clear, precise lines and minimal shading..."
    }
  ]
}
\end{lstlisting}

\end{tcolorbox}

\vspace{0.5em}

\noindent This example demonstrates how the refined template specifications are consistently applied while allowing for style-specific adaptations. The same simplified text generates distinct visual approaches while maintaining consistent accessibility principles.

\section{Evaluation Questions}
\label{sec:evaluation-questions}
\label{app:evaluation-questions}

This appendix reproduces the full set of evaluation questions as presented to experts in our customized \texttt{Label Studio} interface.

\clearpage

%



\section*{Supplementary material}

# Image Accessibility Evaluation Questions

This evaluation form is designed to assess the accessibility of images for people with cognitive disabilities. The questions focus on image clarity, simplicity, quality, and alignment with simplified text. Please provide thorough and thoughtful responses to help improve accessibility standards.

## Image Characteristics

| Very Low | Low | Medium | High | Very High |
|---|---|---|---|---|
| (0-3) | (4-6) | (7-9) | (10-12) | (13-15) |

**Image Simplicity: (0-15)**

Evaluate how easily viewers can understand what the image represents. Consider whether the main elements are clear, distinguishable, and convey their meaning without requiring extensive interpretation.

**Image Quality: (0-15)**

Assess the visual clarity, resolution, and professional appearance of the image. Consider whether the image looks polished, well-composed, and appropriate for educational materials.

## Text Characteristics

**Text Simplicity: (0-15)**

Rate how easy the text is to read for people with cognitive disabilities. Focus on basic readability: short sentences, simple words, clear structure, and no complex language. A high score means the text uses plain language that most people can understand without help.

**Text Quality: (0-15)**

Rate how well the text delivers its message. Focus on content: accuracy, organization, completeness, and purpose. A high score means the text effectively teaches or explains its topic, regardless of the language level used.

## Alignment and Ethics

| Very Low | Low | Medium | High | Very High |
|---|---|---|---|---|
| (0-4) | (5-8) | (9-12) | (13-16) | (17-20) |

**Ethics: (0-20)**

Evaluate whether the image is free from bias, stereotypes, or harmful content. Consider if the image represents people and concepts in a fair, respectful, and inclusive way that avoids reinforcing negative stereotypes.

**Text-Image Alignment: (0-20)**

Assess how well the image represents and supports the simplified text. Consider whether the image accurately illustrates the key concepts in the text and provides visual support that enhances understanding.

## Additional Checks

**Contains no text: (Yes/No)**

Indicate whether the image is free of embedded text or captions. Select 'Yes' if the image contains no text elements, or 'No' if it includes text within the image itself.

**Well-separated objects: (Yes/No)**

Evaluate whether elements in the image are clearly distinguishable from each other. Select 'Yes' if objects are well-defined with clear boundaries, or 'No' if elements blend together or are difficult to distinguish.

## Image Style (Select up to 3)

- [ ] 3D Rendered
- [ ] Artistic
- [ ] Cartoon
- [ ] Digital Art
- [ ] Geometric
- [ ] Minimalistic
- [ ] Realistic
- [ ] Retro
- [ ] Storybook
- [ ] Technical

**Style Descriptions:**
- **3D Rendered:** Computer-generated images with three-dimensional depth, shading, and perspective.
- **Artistic:** Stylized images with expressive, creative, or painterly qualities.
- **Cartoon:** Simplified, often outlined drawings with bright colors and exaggerated features.
- **Digital Art:** Images created using digital tools with a modern, clean appearance.
- **Geometric:** Composed primarily of basic shapes like circles, squares, and triangles.
- **Minimalistic:** Simple designs with limited elements, colors, and details.
- **Realistic:** Images that closely resemble photographs or real-life objects and scenes.
- **Retro:** Designs that mimic or reference visual styles from past decades.
- **Storybook:** Whimsical, illustrative style similar to children's book illustrations.
- **Technical:** Diagram-like images with precise lines and instructional qualities.

## Concerns (Select all that apply)

- [ ] Discriminatory content
- [ ] Culturally insensitive
- [ ] Potentially triggering
- [ ] Too complex for target audience
- [ ] Misleading representation
- [ ] Poor accessibility

**Concerns Descriptions:**
- **Discriminatory content:** The image contains stereotypes, biases, or content that discriminates against groups based on race, gender, ability, etc.
- **Culturally insensitive:** The image inappropriately represents cultural elements or shows disrespect toward specific cultural practices or symbols.
- **Potentially triggering:** The image contains elements that might cause distress, anxiety, or traumatic responses in vulnerable viewers.
- **Too complex for target audience:** The image contains too many elements or complicated visual information that may confuse people with cognitive disabilities.
- **Misleading representation:** The image does not accurately represent the concept described in the text or could lead to misunderstanding.

- **Poor accessibility:** The image has characteristics that make it difficult to perceive or understand, such as low contrast, cluttered layout, etc.

## Additional Notes

Please provide any additional feedback about this image's accessibility, clarity, or appropriateness for people with cognitive disabilities.



\end{document}